\documentclass[preprint,12pt,authoryear]{elsarticle}

\usepackage{amssymb}

\usepackage{amsmath}
\usepackage{graphicx} % Required for inserting images
\usepackage{booktabs}
\usepackage{url}
\usepackage{multirow}
\usepackage{algorithm}
\usepackage{algpseudocode}
\usepackage{pgfplots}
\begin{document}

\begin{frontmatter}

\title{\small{The views expressed are those of the authors and do not reflect the official policy or position of the U.S. Department of War or the U.S. Government.}
\\ 
\vspace{0.2in} 
\Large{Enhancing Heat Sink Efficiency in MOSFETs using Physics Informed Neural Networks: \newline \textit{A Systematic Study on Coolant Velocity Estimation}}} %% Article title

%\title{Enhancing Heat Sink Efficiency in MOSFETs using Physics-Informed Neural Networks: \newline \textit{A Systematic Study on Coolant Velocity Estimation}} %% Article title


%\author{} %% Author name

\author[1]{Aniruddha Bora\corref{cor1}}
\cortext[cor1]{Corresponding author. Email address: aniruddha\_bora@brown.edu; aniruddhabora@gmail.com}
%\ead{aniruddha\_bora@brown.edu}
\author[2]{Isabel K. Alvarez}
\author[2]{Julie Chalfant}
\author[2]{Chryssostomos Chryssostomidis}

\affiliation[1]{Division of Applied Mathematics, Brown University, Providence, RI, USA}
\affiliation[2]{MIT Sea Grant Design Laboratory, Massachusetts Institute of Technology, Cambridge, MA, USA}

\begin{abstract}
In this work, we present a methodology using Physics Informed Neural Networks (PINNs) to determine the required velocity of a coolant, given inlet and outlet temperatures for a given heat flux in a multilayered metal-oxide-semiconductor field-effect transistor (MOSFET). MOSFETs are integral components of Power Electronic Building Blocks (PEBBs) and experiences the majority of the thermal load. Effective cooling of MOSFETs is therefore essential to prevent overheating and potential burnout. Determining the required velocity for the purpose of effective cooling is of importance but is an ill-posed inverse problem and difficult to solve using traditional methods. MOSFET consists of multiple layers with different thermal conductivities, including aluminum, pyrolytic graphite sheets (PGS), and stainless steel pipes containing flowing water. We propose an algorithm that employs sequential training of the MOSFET layers in PINNs. Mathematically, the sequential training method decouples the optimization of each layer by treating the parameters of other layers as constants during its training phase. This reduces the dimensionality of the optimization landscape, making it easier to find the global minimum for each layer's parameters and avoid poor local minima. Convergence of the PINNs solution to the analytical solution is theoretically analyzed. Finally we show the prediction of our proposed methodology to be in good agreement with experimental results.  
\end{abstract}

%% Keywords

\end{frontmatter}

\section{Introduction}
\label{intro}
%%%%%%%%%%%%%%%%%%% MOSFET and HEAT Sink %%%%%%%%%%%%%%

%%% overall goal
Ill-posed problems are ones that do not satisfy Hadamard's conditions for well-posedness: existence, uniqueness, and stability of the solution \cite{hadamard2014lectures}. Conventional numerical methods like finite-element methods or finite-difference methods are robust and effective in solving well-posed problems with complete knowledge of the initial and boundary conditions as well as all material parameters. In practical applications, unfortunately, there are always missing pieces of data and it is necessary to make certain assumptions, e.g. thermal boundary conditions in power electronics cooling applications \cite{toscano2024pinns}. 
For a given setup in which forced convection is used for cooling, the main challenge is to determine the ideal coolant temperature and velocity for a given set of prescribed conditions. This is an ill-posed inverse problem, where multiple solutions may exist with small variations leading to significant different outcomes. Methods like Physics-Informed Neural Networks (PINNs) \cite{raissi2019physics} were developed primarily for such situations, where there is some knowledge of the governing physical laws but not complete knowledge, and there exist some sparse measurements for some of the state variables. As we move towards the era of artificial intelligence, PINNs provide a unique framework that encode physical laws in neural networks and resolve the disconnect between traditional mathematical methods and modern purely data-driven methods \cite{toscano2024pinns}. In heat transfer problems, PINNs are proving to be transformative in both simulation and optimization settings \cite{cai2021physics,bora2021neural,bora2022neural,laubscher2021simulation,xu2023physics,jalili2024physics,zobeiry2021physics,zhang2022pinn,hashemi2024physics,hennigh2021nvidia,norouzi2025cooperative,maryam2023real,ghafoori2025novel}.  The ability to learn from data and identify patterns while constrained in terms of physical laws makes machine learning models like PINNs effective tools.  PINNs with sparse measurements have been shown to generalize well across the whole domain for different fields such as velocity and temperature, even with unknown boundary conditions \cite{cai2021physics,cai2020heat}. Hence, they can be trained on a range of conditions to predict the most effective cooling parameters like water temperature and velocity to maintain the optimal operational environment for maximum efficiency and device performance. 

Due to their efficiency in signal amplification and switching, MOSFETs (Metal-Oxide-Semiconductor Field-Effect Transistors) \cite{barkhordarian1996power} are used in a wide array of electronic devices; however, excessive heat load can lead to reduced efficiency and lifespan of these components \cite{gorecki2022influence,sarkany2016temperature}. There are different approaches to thermal management of MOSFETs. Passive methods are widely used since there is no power consumption and low noise generation \cite{kim2022enhanced}. Passive heat dissipation is enhanced by increasing surface area and reducing thermal interface resistance through the use of thermal interface materials (TIMs) \cite{narumanchi2008thermal,xing2022recent} to facilitate heat dissipation. Forced convection using different fluids is also widely used for accelerating heat dissipation, particularly in high-power applications where passive cooling is not sufficient \cite{purusothaman2018investigation,gorecki2022cooling,kang2012advanced,jorg2017direct}. For applications with very high heat fluxes, integrating phase changing materials (PCMs) into conventional heat sinks has been highlighted as an effective cooling procedure \cite{lu2000thermal,wu2016experimental}. Micro-scale cooling systems are increasing in popularity due to their high heat-flux capacity without the dry-out problems found in PCMs \cite{wei2004stacked,garimella2006chip}. Further comparisons of various heat flux removal techniques can be found in reviews on cooling technologies reported in the literature, e.g. \cite{agostini2007state,ebadian2011review,murshed2017critical}.

%%%%%%%%%%%%%%%%%%%%%%%%%%%%%% iPebbs and Cooling %%%%%%%%%%%%%%%%%%%%%%%%%%\
The idea of electrifying marine vessels has held appeal for both environmental and strategic reasons \cite{IMO2018Strategy}.  The Navy Integrated Power and Energy Corridor (NiPEC) in a shipboard power and energy distribution network is a modular system designed to encompass all aspects of power management, including transmission, conversion, protection, isolation, control, and storage \cite{padilla2021preliminary,cooke2017modular}. Within the NiPEC, Navy integrated Power Electronics Building Blocks (iPEBBs) \cite{rajagopal2021design} are  modular, lightweight, self-contained base units for electrical conversion that NiPEC utilizes to build out safe and resilient high-power shipboard energy systems. Due to the design constraints, conventional thermal management techniques such as direct liquid cooling and finned air-to-air heat exchangers are unsuitable for an iPEBB. Each unit must adhere to strict criteria such as compactness, light weight (less than 35 pounds) and no liquid connections~\cite{alvarez2024experimental,guo2023all}. 

One cooling solution under investigation is indirect liquid cooling \cite{hernandez2023thermal,padilla2023characterizing} in which rack-mounted cold plates are forced into contact with the top and bottom surfaces of the iPEBBs for cooling, with Pyrolytic Graphite Sheets (PGS) used as a thermal interface material. Since iPEBBs are still under development, an adapted version was used for thermal experimentation~\cite{padilla2023characterizing}. The adapted setup has an aluminum cold plate, two additional aluminum plates with pyrolytic graphite sheets (PGS) as thermal interfaces, and four power resistors to generate heat. 

In this paper we propose a new formulation and training strategy for PINNs to determine the effective velocity for indirect dry-interface liquid cooling for power electronics. In our method we define the heat transfer coefficient of the cooling liquid as a parameter of the neural network. In other words, we are solving both the forward problem to determine the temperature distribution as well as the inverse problem to determine the heat transfer coefficient. We introduce a layer-wise training strategy that enhances the accuracy and effectiveness of our approach in addressing these dual objectives by reducing the dimensionality of the optimization landscape. Once the heat transfer coefficient is determined, we utilize the principle of energy conservation to determine the corresponding velocity. 

The rest of the article is organized as follows. %In Section~\ref{sec:background}, we provide background information on neural networks.  
In Section~\ref{sec:methodology}, we introduce the problem setup, the PINNs method, and the algorithm for solving the steady state heat equation for predicting the effective heat transfer coefficient and determining the coolant velocity. In Section~\ref{sec:convergence}, we theoretically analyze the convergence of the proposed solution to the analytical solution. In Section~\ref{sec:numerical}, we test our method in two setups: (i) a toy problem with analytical solution for the heat transfer coefficient and (ii) an experimental setup \cite{alvarez2024experimental}, which is run multiple times with different conditions to compare against PINNs predictions.

\subsection*{Physics Informed Neural Networks (PINNs)}

\begin{figure}[tb]
    \centering
    \includegraphics[width=0.6\textwidth]{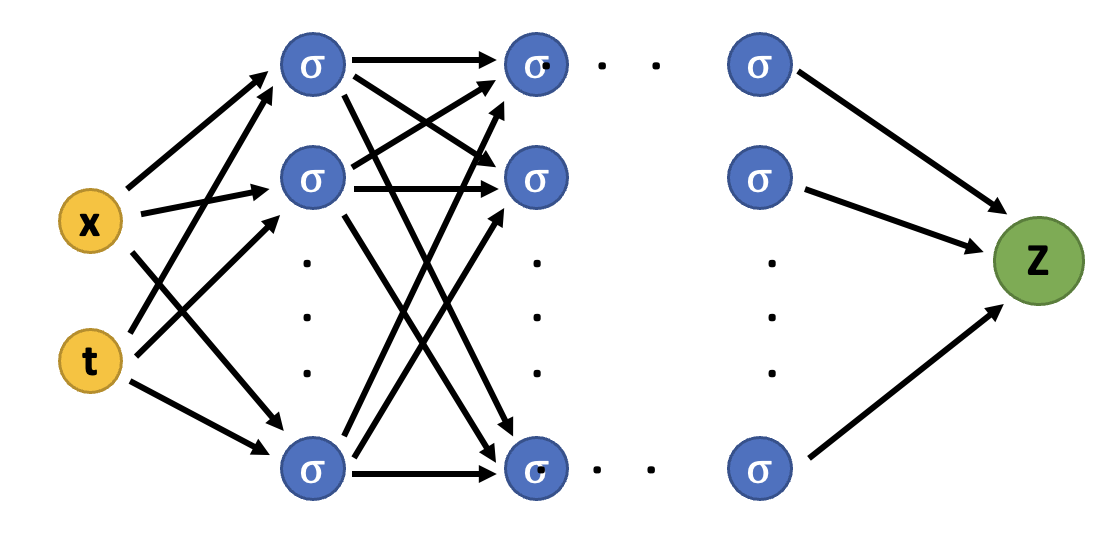}
    \caption{Schematic fully connected neural network  used to solve for the heat equation solution, using PINNs.}
    \label{fig:eg.pinns}
\end{figure}

\begin{figure}[tb]
    \centering
    \includegraphics[width=0.8\textwidth]{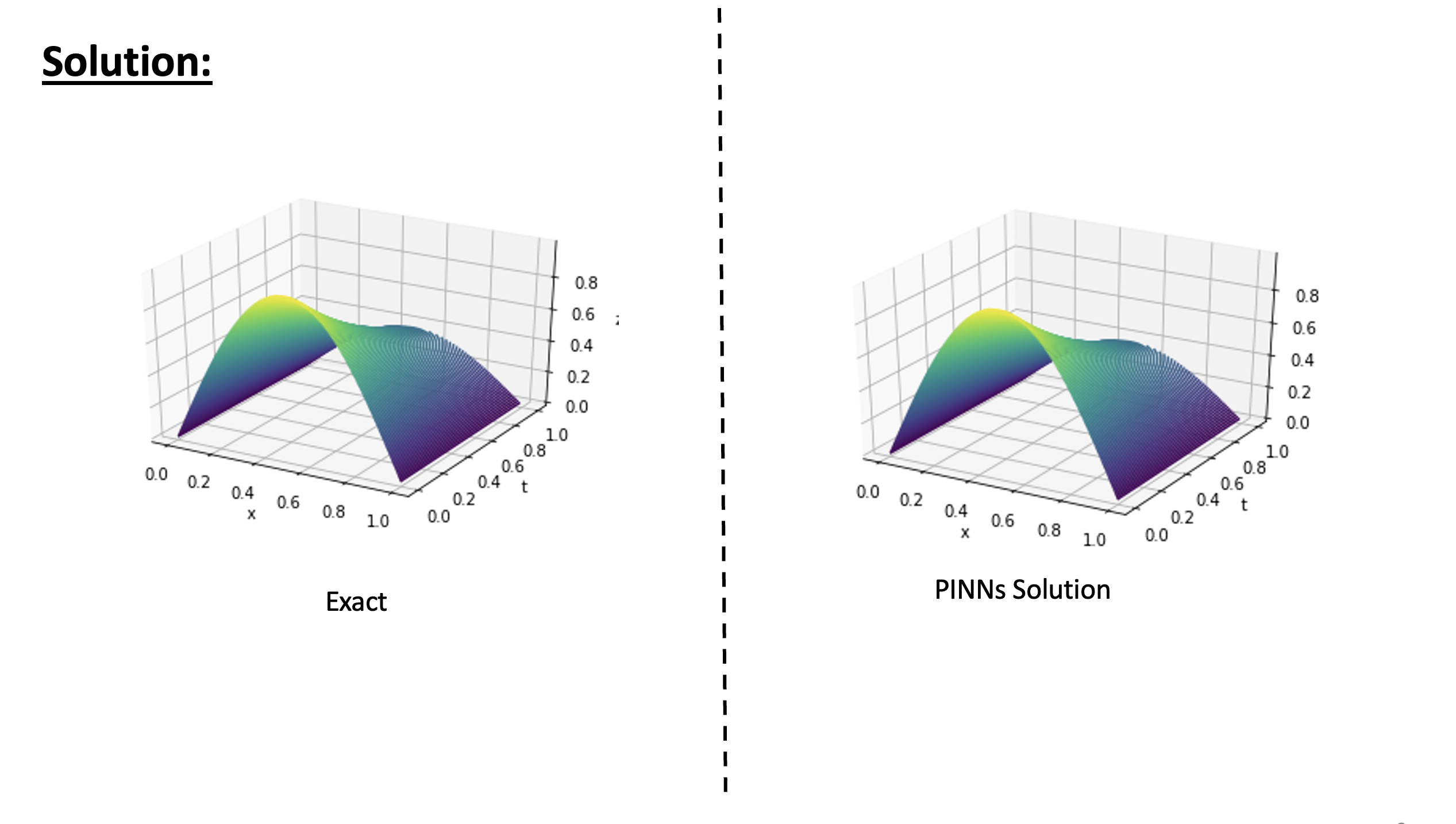}
    \caption{Solution of 1-D transient heat equation}
    \label{fig:1111}
\end{figure}

As stated earlier, Physics Informed Neural Networks (PINNs) encode the laws of physics into the neural networks, thus including them in the training process.  As an example, let us consider a simple heat equation
$$
u_t=u_{x x}+f(x, t), \quad 0<\mathrm{x}<1, \quad 0<\mathrm{t}<1
$$
where,
$$
f(x, t)=e^{-t}\left(1+\pi^2\right) \sin (\pi x),
$$
With the initial conditions:
$$
u(x, 0)=\sin (\pi x), \quad 0<x<1
$$
And boundary conditions:
$$
u(0, t)=u(1, t)=0, \quad 0<t<1
$$
The exact solution is given by: $e^{-t} \sin (\pi x)$. Now we design a fully connected neural network whose output is given by $Z$ as shown in Figure~\ref{fig:eg.pinns}.
We use the following loss function in which
\begin{align}
\text{Loss} &= \frac{1}{N_t N_x} \sum_{i=1}^{N_x} \sum_{j=1}^{N_t} \left(Z_t[i, j] - Z_{xx}[i, j] - f(x_i, t_j)\right)^2 \\
 & + \frac{1}{N_x} \sum_{i=1}^{N_x} \left(Z(x_i, 0) - \sin(\pi x_i)\right)^2 \notag \\
& + \frac{1}{N_t} \sum_{j=1}^{N_t} \left(Z(0, t_j) - 0\right)^2 + \frac{1}{N_t} \sum_{j=1}^{N_t} \left(Z(1, t_j) - 0\right)^2,
\end{align}
where $Z_{t}$, $Z_{xx}$ are the first and second derivatives of $Z$ with respect to $t$ and $x$ respectively calculated by automatic differentiation.  Ideally, as the loss approaches $0$, the output $Z(x,t)$ provided by the neural network approaches $u(x,t)$, which is the solution to the heat equation as defined above. Let us assume $\theta$ to be the collection of all the parameters of the neural network and $\mathcal{L}$ to be the loss. Then the parameters of the neural network are updated based on $\mathcal{L}$ using standard optimizations like gradient descent.  This process is summarized as follows:
\begin{itemize}
    \item[i.] Calculate $\mathcal{L}$.
    \item[ii.] Calculate $\frac{\delta \mathcal{L}}{\delta \theta}$.
    \item[iii.] $\theta_{i}=\theta_{i-1}-\alpha \frac{\delta \mathcal{L}}{\delta \theta}$.
    \item[iv.] if $\mathcal{L} \leq \varepsilon$ or in i $=$ N Stop, else Go to 1.
\end{itemize}
Figure~\ref{fig:1111} shows the exact and the PINNs solutions. The MSE error between the exact and the predicted solution is $5E-08$.

\section{Methodology}
\label{sec:methodology}

\subsubsection*{Problem Description}

\begin{figure}[!ht]
    \centering
    \includegraphics[width=\textwidth]{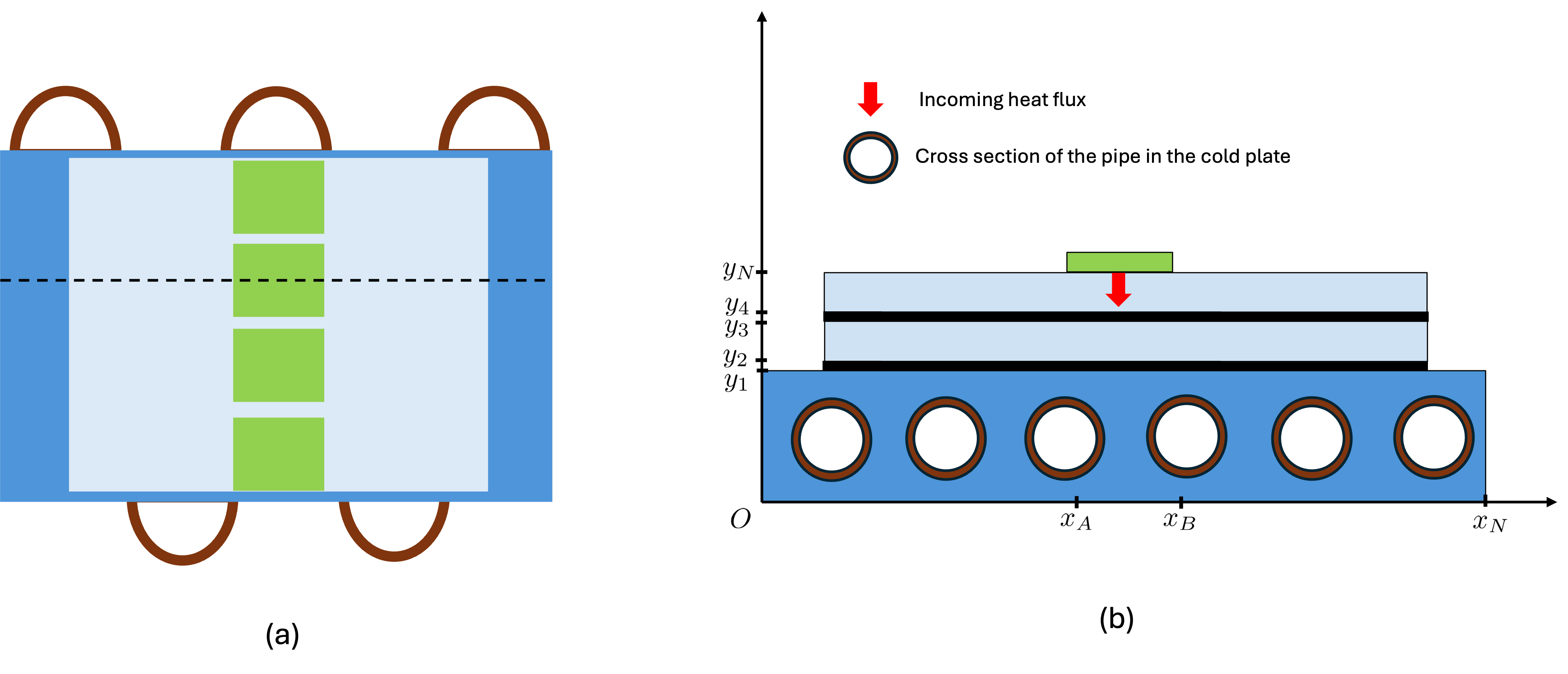}
    \caption{Schematic diagram showing (a) the experimental rig with four power resistors (shown in green) adhered to a layered substrate which is in thermal contact with a cold plate; (b) a cross section of the resistor (shown as dashed line in the left), which is the computational domain for the proposed PINNs.}
    \label{fig:1}
\end{figure}

The experimental setup to be assessed using PINNs is shown in Figure~\ref{fig:1}.  Heat is generated by a set of four resistors, shown in green, aligned to mimic a row of MOSFETs in a PEBB. The resistors are mounted directly onto an aluminum plate using thermal paste.  To imitate a layered substrate and thermal interfaces in the PEBB cooling design, three additional layers are provided: one Pyrolytic Graphite Sheet (PGS), one aluminum plate, and a second PGS layer, followed in sequence by an aluminum cold plate with chilled water flowing through the embedded tubing.  The PINNs model aims to solve the thermal problem at the two-dimensional cross section depicted by a dashed line in the left-hand image in Figure~\ref{fig:1}.

\subsection*{Problem setup}

The following problem setup addresses the geometry presented in Figure~\ref{fig:1}(b).  The layers, designated with an $i$, are numbered as follows:  $i=0$ represents the aluminum cold plate with thermal conductivity of 200 $W/mK$; $i=1$ and $i=3$ represent the PGS layers with thermal conductivity of 0.842 $W/mK$; and $i = 2$ and $i=4$ represent the middle and top aluminum layers with thermal conductivity of 142 $W/mK$. The pipe passes through the cold plate six times and the corresponding cross section of each pass is denoted by $p1$, $p2$, $\cdot \cdot \cdot$, $p6$. The pipe carrying the coolant liquid is made of stainless steel and has a thermal conductivity of 16.2 W/mK. 
Consider the steady state heat conduction equation given by
\begin{align}
    k_{i}(u^{i}_{xx}+u^{i}_{yy}) = 0, \qquad 0 \leq x \leq x_{N}, \quad 0 \leq y \leq y_{N},
\end{align} where $u^{i}$ and $k_{i}$ are the temperature and thermal conductivity for the $i^{th}$ layer, respectively, with periodic boundary conditions for all the layers on the left and right side,
\begin{align}
    \frac{\partial u^{i}(0,y)}{\partial x} &= \frac{\partial u^{i}(x_{N},y)}{\partial x}, \qquad  0 \leq y \leq y_{N}, \quad i=0,1,2,3,4
\end{align} insulated boundary conditions for the top of the top aluminum layer and the bottom of the cold plate
\begin{align}
    \frac{\partial u^{0}(x,0)}{\partial y} &= 0, \qquad 0 \leq x \leq x_{N} , \quad y = 0,\\
    \frac{\partial u^{4}(x,y_{L})}{\partial y} &= 0, \qquad 0 \leq x \leq x_{A}, \quad y= y_{N},   \\
    \frac{\partial u^{4}(x,y_{L})}{\partial y} &= 0, \qquad x_{B} \leq x \leq x_{N}, \quad y= y_{N},
\end{align} Neuman boundary condition for the interface between the MOSFET and the top aluminum plate specifying the heat flux into the system
\begin{align}
     \frac{\partial u(x,y_{L})}{\partial y} &= \alpha,  \qquad x_{A} \leq x \leq x_{B}, \quad y= y_{N},
\end{align}perfectly thermal interface conditions which imply that there is no heat lost when it moves from one layer to the next,
\begin{align}
    u^{i}(x,y) &=  u^{j}(x,y), \qquad (x,y) \in \text{interface} \\
    k_{i}\frac{\partial u^{i}(x,y)}{\partial n} 
    &=  k_{j}\frac{\partial u^{j}(x,y)}{\partial n}, \qquad (x,y) \in \text{interface}
\end{align}and the convective boundary condition from the pipes into the liquid
\begin{align}
    -k_{i}\frac{\partial u^{i}(x,y)}{\partial n} &= h(u^{i}-u^{j}), \quad (x,y) \in \text{inner surface of pipes}
\end{align} where $h$ is the heat transfer coefficient between the stainless steel layer and the coolant fluid, $n$ is the normal to the surface, and $\alpha$ is the flux per unit depth, which is calculated from the flux per unit area, $\beta$,  given by $q^{''}=\beta$. Hence, the flux per unit depth is given by $\alpha = q^{'} = q^{''}/k_{4}$. 

We have two different surface types in our geometry: (i) planar and (ii) circular. For the planar surfaces, we have $\partial/\partial n = \partial/\partial y$. Hence, the interface conditions for the planar regions between layer $i$ and layer $j$ are given by
\begin{align}
    u^{i}(x,y) &=  u^{j}(x,y), \qquad (x,y) \in \text{planar interface} \\
    k_{i}\frac{\partial u^{i}(x,y)}{\partial y} 
    &=  k_{j}\frac{\partial u^{j}(x,y)}{\partial y}, \qquad (x,y) \in \text{planar interface}
\end{align}
Now to derive the normal vector to a circular surface, let us take the general equation of circle:
\begin{equation}
    (x-x_{c})^{2}+(y-y_{c})^{2} = r^{2},
\end{equation}where $(x_{c},y_{c})$ and $r$ are the center and radius of the circle, respectively. Then, 
\begin{align}
    \frac{\partial u}{\partial n} & = \nabla u\cdot n \\
    & = \left(\frac{\partial u}{\partial x}\hat{i} , \frac{\partial u}{\partial y}\hat{j}\right)\cdot\left(\frac{x-x_{c}}{r} \hat{i} , \frac{y-y_{c}}{r}\hat{j}\right)  \quad (x,y) \in \text{circular surface} \\
    & = \frac{\partial u}{\partial x}\left(\frac{x-x_{c}}{r}\right) + \frac{\partial u}{\partial y}\left(\frac{y-y_{c}}{r}\right). \quad (x,y) \in \text{circular surface}
\end{align}
Therefore, for circular interfaces ($i,j$),
\begin{align}
    & u^{i}(x,y) =  u^{j}(x,y) \\
    & k_{i}\left[\frac{\partial u^{i}(x,y)}{\partial x}\left(\frac{x-x_{c}}{r}\right) + \frac{\partial u^{i}(x,y)}{\partial y}\left(\frac{y-y_{c}}{r}\right)\right]  \\
    & = k_{j}\left[\frac{\partial u^{j}(x,y)}{\partial x}\left(\frac{x-x_{c}}{r}\right) + \frac{\partial u^{j}(x,y)}{\partial y}\left(\frac{y-y_{c}}{r}\right)\right],
\end{align}where $(x,y) \in \text{circular interface}$ between the aluminum cold plate and the stainless steel pipes.
%\begin{itemize}
    
%\end{itemize}
\subsection*{Non-dimensionlization}
In order for the neural nets to learn better, we non-dimensionlize the equations. This is important because if the scales of the input parameters/equations are too small or too large the network will either collapse (small values of network parameters) or explode (large value of network parameters). This behavior can be clearly seen from Eqns. (1-5).  We make the above equations dimensionless by introducing the following variables:
\begin{equation}
    x^{\ast} = \frac{x}{x_{L}}, \quad y^{\ast} = \frac{y}{y_{L}}
\end{equation} and
\begin{equation}
    u^{\ast} = \frac{u-U_{0}}{U_{0}},
\end{equation} where $U_{0}$ is the temperature used for normalizing. This temperature is chosen based on the expected scale of the problem, and is usually selected to be the minimum temperature in the system. Then eqn. (1) is given by
\begin{equation}
    k^{\ast}\left(y_{L}^{2}\frac{\partial^{2} u^{\ast}}{\partial x^{\ast 2}}+x_{L}^{2}\frac{\partial^{2} u^{\ast}}{\partial y^{\ast 2}} \right)=0,
\end{equation}where $k^{\ast} = kU_{0}/(x_{L}^{2}y_{L}^{2})$.
The corresponding boundary conditions are given by
\begin{align}
    &\frac{\partial u^{\ast i}(0,y^{\ast})}{\partial x^{\ast}} = \frac{\partial u^{\ast i}(x_{N}/x_{L},y^{\ast})}{\partial x^{\ast}}, \quad 0 \leq y^{\ast} \leq y_{N}/y_{L} \quad i = 0,1,2,3,4 \\
    &\frac{\partial u^{\ast 0}(x^{\ast},0)}{\partial y^{\ast}} = 0, \qquad 0 \leq x^{\ast} \leq x_{N}/x_{L}, \quad y^{\ast} = 0 \\
    &\frac{\partial u^{\ast 4}(x^{\ast},y_{N}/y_{L})}{\partial y^{\ast}} = 0, \qquad 0 \leq x^{\ast} \leq x_{A}/x_{L} \quad y^{\ast} = y_{N}/y_{L}\\
    &\frac{\partial u^{\ast 4}(x^{\ast},y_{N}/y_{L})}{\partial y^{\ast}} = 0, \qquad x_{B}/x_{L} \leq x^{\ast} \leq x_{N}/x_{L} \\
    &\frac{\partial u^{\ast 4}(x^{\ast},y_{N}/y_{L})}{\partial y^{\ast}} = \alpha^{\ast}, \qquad x_{A}/x_{L} \leq x^{\ast} \leq x_{B}/x_{L} \\
    &-\hat{k^{\ast}}_{i}\left(\frac{\partial u^{\ast i}}{\partial x^{\ast}}(\frac{x^{\ast}-x^{\ast}_{cj}}{r^{\ast}})+ \frac{\partial u^{\ast i}}{\partial y^{\ast}}(\frac{y^{\ast}-y^{\ast}_{cj}}{r^{\ast}})\right) = h^{\ast}(u^{\ast}_{i}-u^{\ast}_{ j}), \\
    &\quad (x^{\ast},y^{\ast}) \in  \text{inner surface of pipes}
\end{align}where $i=p1, p2,\cdot \cdot \cdot p6$;  $j=w1, w2,\cdot \cdot \cdot w6$;  $cj$ is the center of the $j^{th}$ circular pipe; $r^{\ast}=r/y_{L}$; $\alpha^{\ast} = \alpha y_{L}/U_{0}$; $h^{\ast}=hU_{0}$ and $\hat{k^{\ast}}=kU_{0}/y_{L}$. The interface conditions in the planar regions
are given by
\begin{align}
    u^{\ast i} &=  u^{\ast j},\\
    \bar{k}_{i}\frac{\partial u^{\ast i}}{\partial y^{\ast}} 
    &=  \hat{k^{\ast}}_{j}\frac{\partial u^{\ast j}}{\partial y^{\ast}},
\end{align}where $\hat{k^{\ast}}_{i}=kU_{0}/y_{L}$.  For the circular interface we have
\begin{align}
    &u^{\ast i} =  u^{\ast j}\\
    &\hat{k}_{i}\left[\frac{\partial u^{\ast}}{\partial x^{\ast}}\left(\frac{x^{\ast}-x^{\ast}_{c}}{r^{\ast}}\right) + \frac{\partial u^{\ast}}{\partial y^{\ast}}\left(\frac{y^{\ast}-y^{\ast}_{c}}{r^{\ast}}\right)\right] \\
    &= \hat{k}_{j}\left[\frac{\partial u^{\ast}}{\partial x^{\ast}}\left(\frac{x^{\ast}-x^{\ast}_{c}}{r^{\ast}}\right) + \frac{\partial u^{\ast}}{\partial y^{\ast}}\left(\frac{y^{\ast}-y^{\ast}_{c}}{r^{\ast}}\right)\right].
\end{align}

\subsection*{Physics Informed Neural Network setup}
\begin{figure}[!ht]
    \centering
    \includegraphics[width=\textwidth]{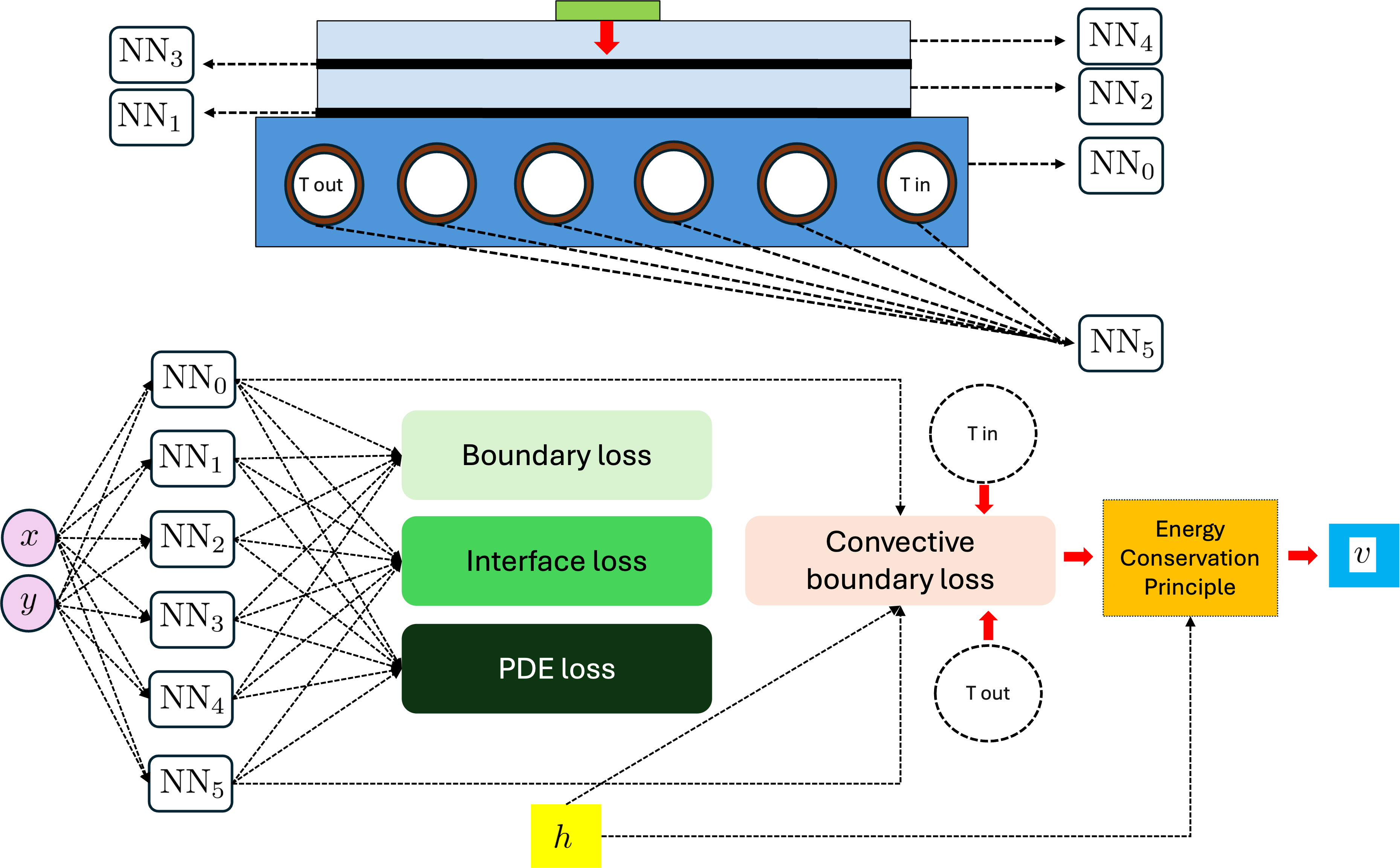}
    \caption{Schematic diagram showing the PINNs setup for the proposed problem statement. Different neural networks are used for different parts of the domain and are united by the loss function. This is done particularly to handle jump discontinuities at the interfaces and scaling variance across different material layers.}
    \label{fig:pinn setup}
\end{figure}

For our problem we chose different neural networks for each layer and combined them using the loss function, eqn (31), as shown in Figure~\ref{fig:pinn setup}. In other words, $NN_{i}$ predicts $u^{i}$, for $i = 0,1,2,3,4$ and $NN_{5}$ predicts $u^{pj}$, for $j = 1,2, \cdot \cdot \cdot, 6$. The loss function combines the results of the individual neural networks together to provide communications between them based on the given conditions. Our setup also includes a trainable parameter $h$, which is the heat transfer coefficient for the coolant fluid (water in our case), predicted based on the optimized loss function. The individual neural networks take in $x$ and $y$ for their respective domains and predict the temperature throughout. Additionally, the loss term for the convective boundary in the inner surface of the pipes takes in the $T_{in}$ and $T_{out}$, which are the inlet and outlet temperatures of the coolant fluid. The energy conservation principle, i.e., the total amount of heat in must equal the total amount of heat out, is imposed in the loss function to guarantee uniqueness. The details of the loss function are discussed in the next section.
\subsubsection*{Loss Function}
The loss function for the proposed method is given by
\begin{align}
\mathcal{L} = \lambda_{0}L_{PDE} + \lambda_{1}L_{BC} + \lambda_{2}L_{IC} + \lambda_{3}L_{CB} + \lambda_{4}L_{h} + \lambda_{5}L_{Q} + \lambda_{6}L_{Data},
\end{align}where $\lambda_{s}|_{s=0}^{6}$ are self adaptive weights \cite{mcclenny2023self,anagnostopoulos2024residual}. The use of self-adaptive weights helps to balance the contribution of the data-driven learning and the physics-based constraints during the training process. These self-adaptive weights are coefficients that adjust automatically during the training to prioritize either the fidelity to the empirical data or the adherence to the physical laws encoded in the model. The individual loss terms in Eqn.~(31) are given by
\begin{align}
    L_{PDE} & = \sum_{i=0}^{4}||k^{i \ast}\left(y_{L}^{2}\frac{\partial^{2} u^{i \ast}}{\partial x^{\ast 2}}+x_{L}^{2}\frac{\partial^{2} u^{i \ast}}{\partial y^{\ast 2}} \right)- 0 ||_{2} + \notag \\
    &\hspace{0.2in} \sum_{i=0}^{6}||k^{i \ast}\left(y_{L}^{2}\frac{\partial^{2} u^{pi \ast}}{\partial x^{\ast 2}}+x_{L}^{2}\frac{\partial^{2} u^{pi \ast}}{\partial y^{\ast 2}} \right)- 0 ||_{2} \\
    L_{BC}  &=  \sum_{i=0}^{4}||\frac{\partial u^{\ast i}(0,y^{\ast})}{\partial x^{\ast}} - \frac{\partial u^{\ast i}(1,y^{\ast})}{\partial x^{\ast}}||_{2}{}_{\hspace{0.05in} y^{\ast} \in [0,y_{N}/y_{L}]} +\notag\\
    & ||\frac{\partial u^{\ast 0}(x^{\ast},0)}{\partial y^{\ast}}-0||_{2}{}_{\hspace{0.05in} x^{\ast} \in [0, x_{N}/x_{L}]} + ||\frac{\partial u^{\ast 2}(x^{\ast},1)}{\partial y^{\ast}} - 0||_{2}{}_{\hspace{0.05in} x^{\ast} \in [0 , x_{A}/x_{L}]}+ \notag\\ 
    & ||\frac{\partial u^{\ast 2}(x^{\ast},1)}{\partial y^{\ast}} - 0||_{2}{}_{\hspace{0.05in} x^{\ast} \in [x_{B}/x_{L}, x_{N}/x_{L}]} + ||\frac{\partial u^{\ast 2}(x^{\ast}_{b},1)}{\partial y^{\ast}} - \alpha^{\ast} ||_{2}{}_{\hspace{0.05in} x^{\ast} \in[x_{A}/x_{L}, x_{B}/x_{L}]}\\
    L_{IC} &=  \sum_{i,j=0,1}^{1,2}||u^{\ast i}(x,y) - u^{\ast j}(x,y)||_{2}+\sum_{i,j=0,1}^{1,2}||\bar{k}_{i}\frac{\partial u^{\ast i}}{\partial y^{\ast}} 
    - \bar{k}_{j}\frac{\partial u^{\ast j}}{\partial y^{\ast}}||_{2} + \notag\\
    & \sum_{i=p1}^{p6}||u^{\ast 0} - u^{\ast j}||_{2} + \sum_{j=p1}^{p6}||\hat{k}_{0}\left[\frac{\partial u^{ 0 \ast}}{\partial x^{\ast}}\left(\frac{x^{\ast}-x^{\ast}_{cj}}{r^{\ast}}\right) + \frac{\partial u^{0 \ast}}{\partial y^{\ast}}\left(\frac{y^{\ast}-y^{\ast}_{cj}}{r^{\ast}}\right)\right] \notag \\
    & - \hat{k}_{j}\left[\frac{\partial u^{j \ast}}{\partial x^{\ast}}\left(\frac{x^{\ast}-x^{\ast}_{cj}}{r^{\ast}}\right) + \frac{\partial u^{j \ast}}{\partial y^{\ast}}\left(\frac{y^{\ast}-y^{\ast}_{c}}{r^{\ast}}\right)\right]||_{2}
\end{align}

\begin{align}
    L_{CB} &= \sum_{i,j=p1,w1}^{p6,w6}||-\hat{k}_{i}\left(\frac{\partial u^{\ast i}}{\partial x^{\ast}}(\frac{x^{\ast}-x^{\ast}_{ci}}{r^{\ast}}\frac{y_{L}}{x_{L}})+ \frac{\partial u^{\ast i}}{\partial y^{\ast}}(\frac{y^{\ast}-y^{\ast}_{ci}}{r^{\ast}})\right) - h^{\ast}(u^{\ast i}-u^{\ast j})||_{2} \\
    L_{h} &=  \sum_{j=p1}^{p6}||-\hat{k}_j(u^{\ast j}_x(x^{\ast}_{int j}-x^{\ast}_{cj})/r^{\ast}+u^{\ast j}_y(y^{\ast}_{intj}-y^{\ast}_{cj})/r^{\ast})-h^{\ast}(u^{\ast j}-t_w^{\ast})||_{2},
\end{align}where $(x_{intj}, y_{intj})$ are the points from the boundary of the pipes $j$, which are in contact with the water. $t_w^{\ast}$ is the dimensionless temperature, which is equal to $(t^{\ast}_{in}+t^{\ast}_{out})/2$, where $t^{\ast}_{in}$ and $t^{\ast}_{out}$ are the dimensionless version of the given water inlet and outlet temperatures, non-dimensionalized using $U_{0}$. 
\begin{align}
    L_{Q} & = ||Q_{in} - Q_{out}||_{2},
\end{align}
where 
\begin{align}
     Q_{in} & = P_{0}\\
    Q_{out} & = hA[\sum_{i=0}^{6}(t_i-t_w)].
\end{align}
Here $Q_{in}$ is the amount of heat coming in to the system and $Q_{out}$ is the amount of heat going out of the system. $A=2\pi rl$, where $r$ is the radius of the pipe, and $l$ is the total length of the pipe in contact with the cold plate. $t_{i}$ is the maximum temperature at the boundary of the $i^{th}$ pipe containing the water. 

\subsubsection*{Calculation of velocity}
To determine the velocity, we use two equations. The first one captures the heat transfer from the pipe to the water
\begin{equation}
    Q = h \cdot A_1 \cdot \sum_{i=p1}^{p6}\Delta T_i/6,
    \label{eq:pipeheat1}
\end{equation}where $h$ is the heat transfer coefficient, $A_1$ is the surface area of the pipes, and $\Delta T_i$ is the temperature difference between the $i^{th}$ pipe's inner surface and temperature of the cooling fluid. The second equation describes the heat convected away by the water flowing through the pipes
\begin{equation}
    Q = \rho \cdot A_2 \cdot v \cdot c_p \cdot \Delta T_2, 
    \label{eq:pipeheat2}
\end{equation}where $\rho$ is the density of the cooling fluid, $A_2$ is the cross-sectional area of the pipes, $v$ the velocity of the water, $c_p$ is the specific heat of water, and $\Delta T_2$ the temperature differential between the outlet and inlet cooling fluid temperature. These equations establish a relationship between the heat transferred from the pipe to the water and the heat carried away by the water. By setting them equal to each other, we assert that the system is in a steady state, with the input and output heat fluxes balanced. To calculate the necessary velocity of the water to achieve this balance, we derive the velocity expression using equations \eqref{eq:pipeheat1} and \eqref{eq:pipeheat2}
\begin{equation}
    v_{\text{NN}} = \frac{h \cdot A_1 \cdot \sum_{i=1}^{6}\Delta T_i/6}{\rho \cdot A_2 \cdot C_p \cdot \Delta T_2}
\end{equation} The subscript ``NN" in $v_{\text{NN}}$ indicates that this velocity is the value that we predict through the machine learning method based on predicted $h$ value. 
\subsection{Layer-wise sequential training}
Training the neural network sequentially, layer by layer, is often more effective than training all layers simultaneously because it simplifies the optimization problem at each stage, leading to improved convergence and stability. By focusing on one layer at a time, the network can accurately learn the specific features and boundary conditions associated with that layer without interference from other layers. This approach ensures that the interface conditions between layers are satisfied more precisely, as each subsequent layer builds upon the already optimized previous layers. Overall, sequential training allows the neural network to model complex multilayer systems more effectively by reducing complexity, enhancing stability, and improving the satisfaction of physical constraints. From a mathematical standpoint, sequential training reduces the complexity of the optimization problem by partitioning it into smaller, more manageable subproblems. Consider the total loss function \( J_{\text{Total}}(\theta) \) for all layers:
\begin{equation}
    J_{\text{Total}}(\theta) = \sum_{i=1}^{N} J^{i}(\theta^{i}, \theta^{<i}),
\end{equation} where \( J^{i} \) is the loss for layer \( i \), \( \theta^{i} \) are the parameters for layer \( i \), and \( \theta^{<i} \) represents the parameters of the previously trained layers. In sequential training, we optimize \( J^{i} \) with respect to \( \theta^{i} \) while treating \( \theta^{<i} \) as constants, effectively decoupling the optimization problem for each layer:
\begin{equation}
    \theta^{i*} = \arg\min_{\theta^{i}} J^{i}(\theta^{i}; \theta^{<i}).
\end{equation} This decoupling simplifies the optimization landscape, making it easier to find the global minimum for each layer's parameters without the interference of parameter interactions present in joint training. Additionally, by training layers sequentially, we ensure that the boundary and interface conditions are satisfied at each step, reducing the cumulative error across layers. Mathematically, the interface conditions between layers \( i \) and \( i+1 \) are enforced by minimizing:
\begin{equation}
    \left\| u_{\text{NN}}^{i}(x, y) - u_{\text{NN}}^{i+1}(x, y) \right\|_{L^{2}(\Gamma_{i, i+1})} \leq \varepsilon,
\end{equation} where \( \Gamma_{i, i+1} \) is the interface between layers \( i \) and \( i+1 \), and \( \varepsilon \) is a small threshold. Sequential training allows for tighter control over this condition, leading to better overall convergence of the neural network to the true solution. The overall steps for the above described methodology are shown in Algorithim 1.
\begin{figure}[!ht]
    \centering
    \includegraphics[width=1.0\linewidth]{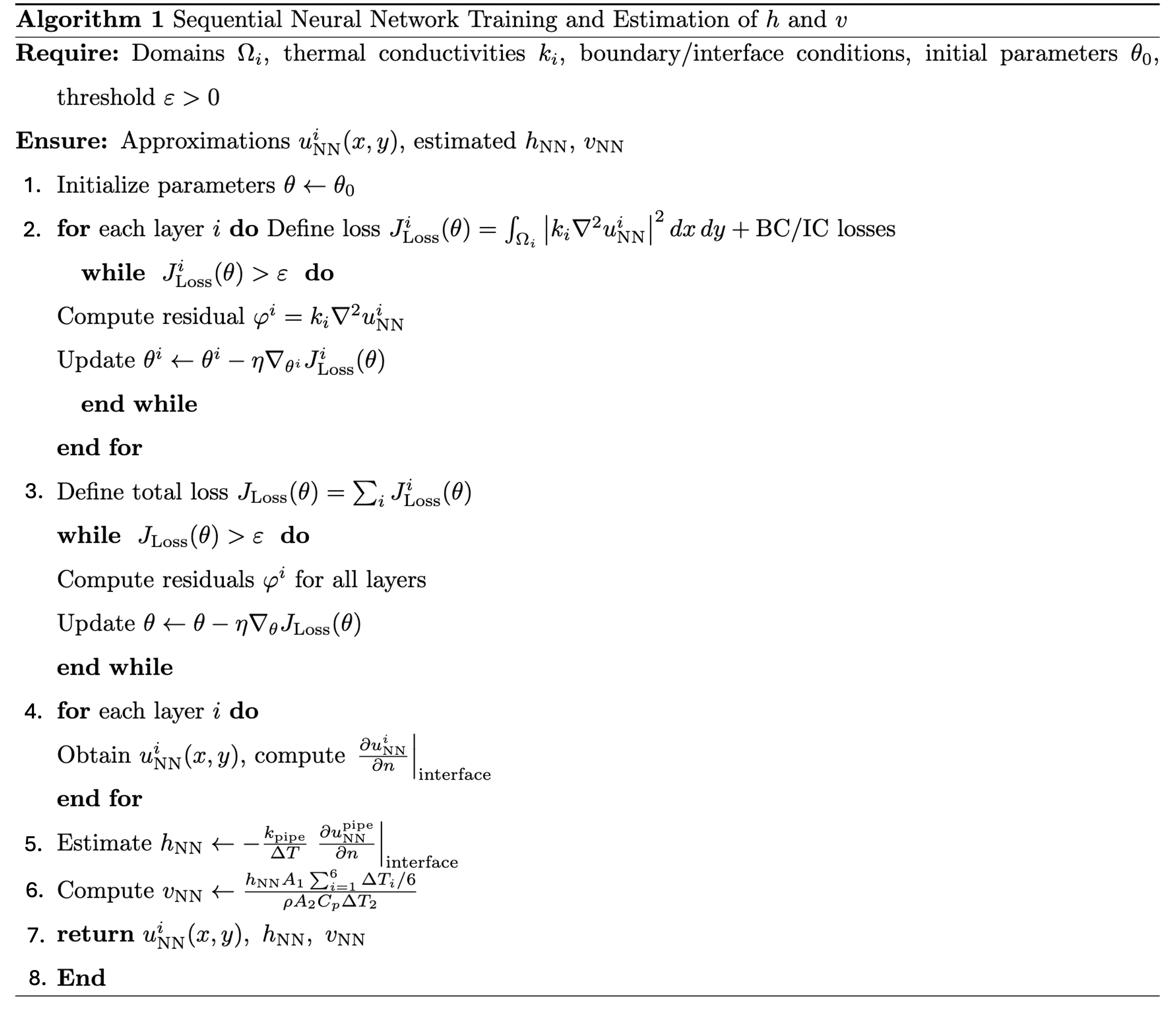}
\end{figure}

\section{Convergence analysis}
\label{sec:convergence}
In this section we provide a theoretical analysis of our proposed methodology, to show that the predicted results converge to the analytical solution for the set of Eqns (8-16).

\textbf{Theorem}. Assume that the analytical solutions $\{u^{i}(x, y)\}$ and the neural network approximations $\{u_{\text{NN}}^{i}(x, y)\}$ are smooth and defined on their respective domains $\Omega_{i}$ for $i = 0, 1, 2, 3, 4,$ 
$\text{ p1, p2, ..., p6,}$ $\text{w1, w2, ..., w6}$. Suppose that the loss function \(J_{\text{Loss}}(\theta) \leq \varepsilon\), where \(\varepsilon\) is a small positive constant. Let the error between the neural network solution and the analytical solution be defined as: \begin{equation}
    E^{i}(x, y) = u_{\text{NN}}^{i}(x, y) - u^{i}(x, y),
\end{equation} then, there exists a constant \(A\) such that:
\begin{equation}
    \sum_{i} \int_{\Omega_{i}} (E^{i})^{2}\, dx\, dy \leq \varepsilon A.
\end{equation} Furthermore, the neural network estimates of the heat transfer coefficient \(h_{\text{NN}}\) and the velocity \(v_{\text{NN}}\) will converge to the true values \(h\) and \(v\) as \(\varepsilon \to 0\).

\textbf{Proof:} 

\textbf{Part A:} The steady-state heat conduction equation for each layer \(i\) is given by:
\begin{equation}
    k_{i} \left( \frac{\partial^{2} u^{i}}{\partial x^{2}} + \frac{\partial^{2} u^{i}}{\partial y^{2}} \right) = 0, \quad \text{in } \Omega_{i}.
\end{equation} The neural network approximations \(u_{\text{NN}}^{i}\) satisfy:
\begin{equation}
    k_{i} \left( \frac{\partial^{2} u_{\text{NN}}^{i}}{\partial x^{2}} + \frac{\partial^{2} u_{\text{NN}}^{i}}{\partial y^{2}} \right) = \varphi^{i}(x, y), \quad \text{in } \Omega_{i},
\end{equation} where \(\varphi^{i}(x, y)\) represents the residual resulting from the neural network approximation, and we assume:
\begin{equation}
    \varphi^{i}\|_{L^{2}(\Omega_{i})} \leq \varepsilon.
\end{equation} Subtracting the analytical PDE from the neural network PDE, we obtain the error equation:
\begin{equation}
    k_{i} \left( \frac{\partial^{2} E^{i}}{\partial x^{2}} + \frac{\partial^{2} E^{i}}{\partial y^{2}} \right) = \varphi^{i}(x, y), \quad \text{in } \Omega_{i}.
\end{equation} The analytical solutions satisfy certain boundary and interface conditions, which the neural network approximations satisfy up to a small error bounded by \(\varepsilon\). Specifically:
\begin{itemize}
    \item Planar Interfaces (between layers \(i\) and \(j\))
  \[
  \begin{aligned}
  E^{i}(x, y) - E^{j}(x, y) &= \eta_{ij}^{1}(x, y), \\
  k_{i} \frac{\partial E^{i}}{\partial y} - k_{j} \frac{\partial E^{j}}{\partial y} &= \eta_{ij}^{2}(x, y),
  \end{aligned}
  \]

  where \(|\eta_{ij}^{1}(x, y)| \leq \varepsilon\) and \(|\eta_{ij}^{2}(x, y)| \leq \varepsilon\).
 \item Circular Interfaces (between layers \(i\) and \(j\))
  \[
  \begin{aligned}
  &E^{i}(x, y) - E^{j}(x, y) = \eta_{ij}^{1}(x, y), \\
  &k_{i} \left( \frac{\partial E^{i}}{\partial x} \frac{x - x_{c}}{r} + \frac{\partial E^{i}}{\partial y} \frac{y - y_{c}}{r} \right) - k_{j} \left( \frac{\partial E^{j}}{\partial x} \frac{x - x_{c}}{r} + \frac{\partial E^{j}}{\partial y} \frac{y - y_{c}}{r} \right) \\
  & = \eta_{ij}^{3}(x, y),
  \end{aligned}
  \]

  where \(|\eta_{ij}^{3}(x, y)| \leq \varepsilon\).

\end{itemize}
Multiplying  both sides of the error equation by \(E^{i}\) and integrating over \(\Omega_{i}\):
\begin{equation}
    \int_{\Omega_{i}} k_{i} \left( \frac{\partial^{2} E^{i}}{\partial x^{2}} + \frac{\partial^{2} E^{i}}{\partial y^{2}} \right) E^{i}\, dx\, dy = \int_{\Omega_{i}} \varphi^{i} E^{i}\, dx\, dy.
\end{equation} Using integration by parts:
\begin{equation}
    - k_{i} \int_{\Omega_{i}} |\nabla E^{i}|^{2}\, dx\, dy + k_{i} \int_{\partial \Omega_{i}} \frac{\partial E^{i}}{\partial n} E^{i}\, ds = \int_{\Omega_{i}} \varphi^{i} E^{i}\, dx\, dy.
\end{equation} The boundary integral can be decomposed into contributions from the physical boundaries and the interfaces between layers. Since the errors in the boundary and interface conditions are bounded by \(\varepsilon\), we have:
\begin{equation}
    \left| k_{i} \int_{\partial \Omega_{i}} \frac{\partial E^{i}}{\partial n} E^{i}\, ds \right| \leq \varepsilon B_{1},
\end{equation} where \(B_{1}\) is a constant depending on the maximum values of \(E^{i}\) and \(\frac{\partial E^{i}}{\partial n}\), which are bounded due to the smoothness of \(E^{i}\). Applying the Cauchy-Schwarz inequality:
\begin{equation}
    \left| \int_{\Omega_{i}} \varphi^{i} E^{i}\, dx\, dy \right| \leq \|\varphi^{i}\|_{L^{2}(\Omega_{i})} \|E^{i}\|_{L^{2}(\Omega_{i})} \leq \varepsilon \|E^{i}\|_{L^{2}(\Omega_{i})}.
\end{equation} Summing over all layers \(i\), we have:

\begin{equation}
    - \sum_{i} k_{i} \int_{\Omega_{i}} |\nabla E^{i}|^{2}\, dx\, dy \leq \sum_{i} \varepsilon \|E^{i}\|_{L^{2}(\Omega_{i})} + \varepsilon B_{1}.
\end{equation} Rewriting,
\begin{equation}
    \sum_{i} k_{i} \int_{\Omega_{i}} |\nabla E^{i}|^{2}\, dx\, dy \geq - \left( \sum_{i} \varepsilon \|E^{i}\|_{L^{2}(\Omega_{i})} + \varepsilon B_{1} \right).
\end{equation} Since the left-hand side is non-negative, we have,
\begin{equation}
    0 \geq - \left( \sum_{i} \varepsilon \|E^{i}\|_{L^{2}(\Omega_{i})} + \varepsilon B_{1} \right).
\end{equation} This inequality holds trivially, but to obtain a useful bound, we proceed by estimating \(\|E^{i}\|_{L^{2}(\Omega_{i})}\). Assuming that \(E^{i}\) satisfies appropriate boundary conditions (e.g., zero on a portion of the boundary), the Poincaré inequality applies:
\begin{equation}
    \|E^{i}\|_{L^{2}(\Omega_{i})} \leq C_{P} \|\nabla E^{i}\|_{L^{2}(\Omega_{i})},
\end{equation} where \(C_{P}\) is the Poincaré constant for domain \(\Omega_{i}\). Using the Poincaré inequality:
\begin{equation}
    \sum_{i} k_{i} \int_{\Omega_{i}} |\nabla E^{i}|^{2}\, dx\, dy \geq - \left( \varepsilon C_{P} \sum_{i} \|\nabla E^{i}\|_{L^{2}(\Omega_{i})} + \varepsilon B_{1} \right).
\end{equation} Let \(S = \sum_{i} \|\nabla E^{i}\|_{L^{2}(\Omega_{i})}^{2}\). Then we have,
\begin{equation}
    k_{\min} S \geq - \left( \varepsilon C_{P} \sqrt{N S} + \varepsilon B_{1} \right),
\end{equation} where \(k_{\min} = \min_{i} k_{i}\) and \(N\) is the number of layers. Rewriting it gives us
\begin{equation}
    k_{\min} S + \varepsilon C_{P} \sqrt{N S} + \varepsilon B_{1} \geq 0.
\end{equation} This is a quadratic in \(\sqrt{S}\). Solving for \(S\), we find that \(S\) is bounded. Using the Poincaré inequality again:
\begin{equation}
    \|E^{i}\|_{L^{2}(\Omega_{i})} \leq C_{P} \|\nabla E^{i}\|_{L^{2}(\Omega_{i})} \leq C_{P} \sqrt{S}.
\end{equation} Therefore, the total \(L^{2}\) norm of the error is bounded:
\begin{equation}
    \sum_{i} \|E^{i}\|_{L^{2}(\Omega_{i})}^{2} \leq N C_{P}^{2} S \leq \varepsilon^{2} A,
\end{equation} where \(A\) is a constant depending on \(k_{\min}\), \(C_{P}\), \(B_{1}\), and \(N\).

\textbf{Part B:} The neural network estimates the heat transfer coefficient \(h_{\text{NN}}\) based on the temperature distribution \(u_{\text{NN}}^{i}\). The true \(h\) is determined by the gradient of the true temperature at the interface between the pipes and water:
\begin{equation}
    h = -\frac{k_{\text{pipe}}}{\Delta T} \left. \frac{\partial u^{\text{pipe}}}{\partial n} \right|_{\text{interface}},
\end{equation} where \(\Delta T\) is the temperature difference between the pipe surface and the water. Similarly, the neural network estimate is:
\begin{equation}
    h_{\text{NN}} = -\frac{k_{\text{pipe}}}{\Delta T} \left. \frac{\partial u_{\text{NN}}^{\text{pipe}}}{\partial n} \right|_{\text{interface}}.
\end{equation} The error in \(h\) is:
\begin{equation}
    E_{h} = h_{\text{NN}} - h = -\frac{k_{\text{pipe}}}{\Delta T} \left( \left. \frac{\partial u_{\text{NN}}^{\text{pipe}}}{\partial n} - \frac{\partial u^{\text{pipe}}}{\partial n} \right|_{\text{interface}} \right).
\end{equation} Since \(u_{\text{NN}}^{\text{pipe}} - u^{\text{pipe}} = E^{\text{pipe}}\) and their gradients differ by \(\nabla E^{\text{pipe}}\), which is bounded by \(\sqrt{S}\), we have:
\begin{equation}
    |E_{h}| \leq \frac{k_{\text{pipe}}}{\Delta T} \left| \left. \frac{\partial E^{\text{pipe}}}{\partial n} \right|_{\text{interface}} \right| \leq C \varepsilon,
\end{equation} where \(C\) is a constant depending on the materials and geometry. Similarly, the velocity \(v_{\text{NN}}\) is computed using:
\begin{equation}
    v_{\text{NN}} = \frac{h_{\text{NN}} \cdot A_{1} \cdot \sum_{i=p1}^{p6} \Delta T_{i}/6}{\rho \cdot A_{2} \cdot C_{p} \cdot \Delta T_{2}}.
\end{equation} The error in \(v\) is:
\begin{equation}
    E_{v} = v_{\text{NN}} - v = \frac{E_{h} \cdot A_{1} \cdot \sum_{i=p1}^{p6} \Delta T_{i}/6}{\rho \cdot A_{2} \cdot C_{p} \cdot \Delta T_{2}}.
\end{equation}
Therefore:
\begin{equation}
    |E_{v}| \leq \frac{|E_{h}| \cdot A_{1} \cdot \sum_{i=p1}^{p6} \Delta T_{i}/6}{\rho \cdot A_{2} \cdot C_{p} \cdot \Delta T_{2}} \leq C' \varepsilon,
\end{equation} where \(C'\) is a constant. Thus, both \(h_{\text{NN}}\) and \(v_{\text{NN}}\) converge to the true values \(h\) and \(v\) as \(\varepsilon \to 0\), with errors bounded by \(C \varepsilon\) and \(C' \varepsilon\) respectively.

\section{Numerical Results}
\label{sec:numerical}
In this section we show two different sets of numerical experiments. In the first part we explore a problem for which an analytical result is known, which allows us to validate our formulation and methodology. In the second part we show the applicability of our methodology to a real problem and compare with experimental results to further validate our methodology.
\subsubsection*{Part A: Analytical example}
\begin{figure}[!ht]
    \centering
    \includegraphics[width=0.5\linewidth]{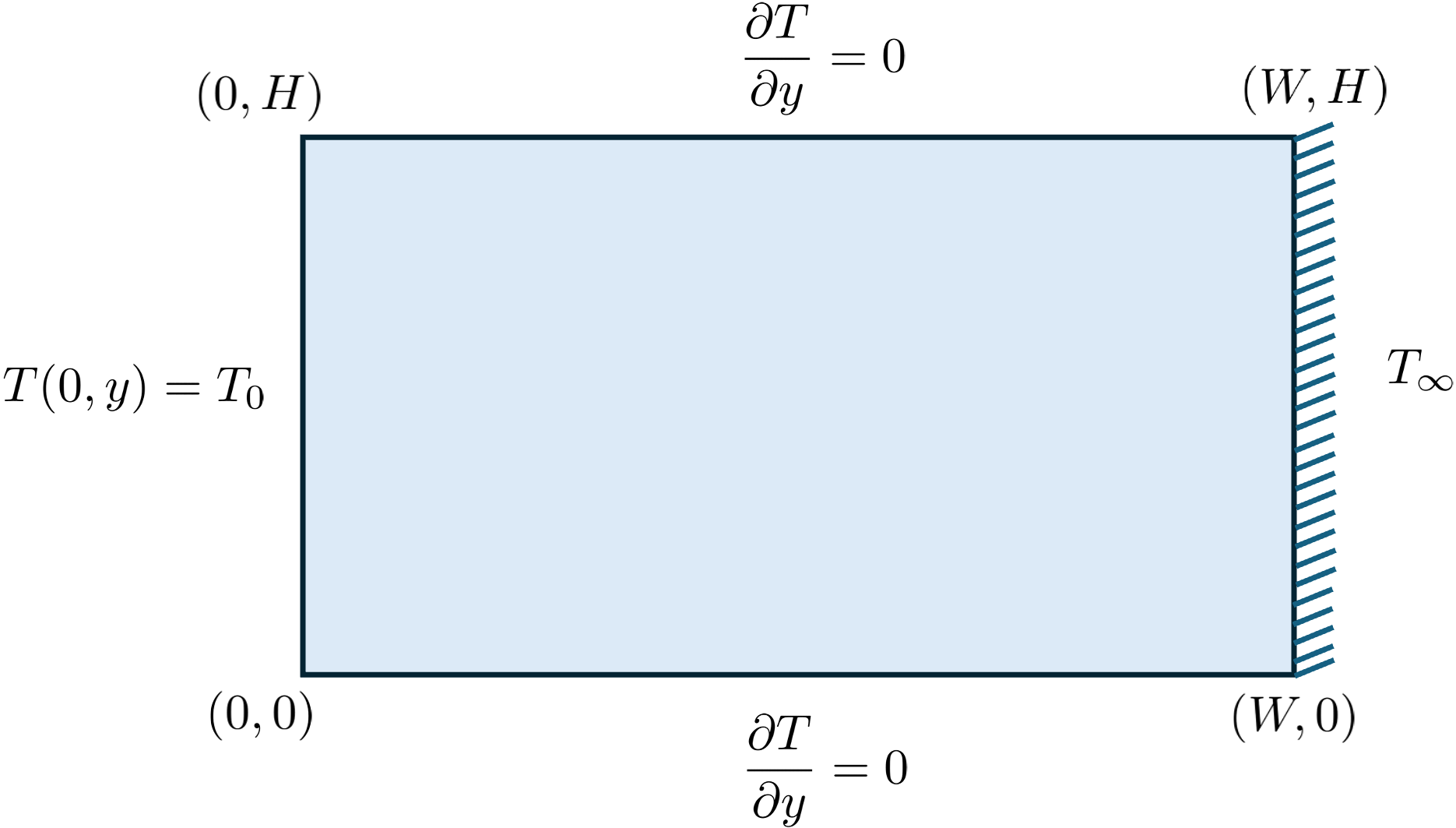}
    \caption{Schematic diagram showing the computational domain of the analytical example.}
    \label{fig:A1a}
\end{figure}
Let us consider the steady-state heat conduction equation in two dimensions
\begin{equation}
\frac{\partial^2 T}{\partial x^2} + \frac{\partial^2 T}{\partial y^2} = 0, \quad \text{for} \ 0 < x < W, \ 0 < y < H
\end{equation}with Dirichlet boundary condition
\begin{equation}
T(0, y) = T_0, \quad \text{for} \ 0 \leq y \leq H
\end{equation}
insulated boundary condition
\begin{equation}
\frac{\partial T}{\partial y} \Bigg|_{y=0} = 0, \quad \frac{\partial T}{\partial y} \Bigg|_{y=H} = 0, \quad \text{for} \ 0 \leq x \leq W
\end{equation}and convective boundary condition
\begin{equation}
-k \frac{\partial T}{\partial x} \Bigg|_{x=W} = h[T(W, y) - T_{\infty}], \quad \text{for} \ 0 \leq y \leq H
\end{equation}where $h$ is the heat transfer coefficient for that medium. The computational domain is shown in Figure~\ref{fig:A1a}. We create this example, since the analytical expression for the heat transfer coefficient can be easily derived and can be used to validate the ability of PINNs to determine the heat transfer coefficient, which we use as a guidance to determine the velocity of the cooling fluid for the experimental setup. Assuming the temperature is independent of \( y \) due to the insulated boundaries (since there is no heat flux in the \( y \)-direction), the problem simplifies to one-dimensional heat conduction in the \( x \)-direction. Thus, the governing equation reduces to:
\begin{equation}
\frac{d^2 T}{dx^2} = 0, \quad \text{for} \ 0 < x < W.
\end{equation} Integrating twice we have
\begin{equation*}
\frac{d^2 T}{dx^2} = 0 \quad \Rightarrow \quad \frac{dT}{dx} = C_1 \quad \Rightarrow \quad T(x) = C_1 x + C_2.
\end{equation*}
Applying the boundary conditions at \( x = 0 \):
\begin{equation}
T(0) = T_0 \quad \Rightarrow \quad C_2 = T_0
\end{equation}and using the convective boundary condition at  \( x = W \)
\begin{equation}
-k \frac{dT}{dx} \Bigg|_{x=W} = h[T(W) - T_{\infty}]
\end{equation}the final temperature distribution can be given by
\begin{equation}
T(x) = \frac{-h(T_0 - T_{\infty})}{k + h W} x + T_0.
\end{equation}
To compute the heat transfer coefficient \( h \), we simulate the temperature at \( x = W \). From the temperature distribution:
\begin{equation}
T(W) = \frac{-h(T_0 - T_{\infty})}{k + h W} W + T_0.
\end{equation}Now solving for \( h \):
\begin{equation}
h = \frac{(T_0 - T(W)) k}{W (T(W) - T_{\infty})}. 
\end{equation}\textbf{Example: } Let the dimensions of the rectangular plate be given as
        \begin{align*}
            W &= 0.1 \, \text{m} \quad (\text{Width of the plate}) \\
            H &= 0.1 \, \text{m} \quad (\text{Height, insulated, so temperature variation in } y \text{ is negligible})
        \end{align*}
with thermal conductivity of $k=20$ $W/mK$ and boundary temperatures
        \begin{align*}
            T_0 &= 100^\circ \text{C} \quad (\text{Left side temperature}) \\
            T_\infty &= 25^\circ \text{C} \quad (\text{Ambient temperature})
        \end{align*}
\begin{figure}[!ht]
    \centering
    \includegraphics[width=\textwidth]{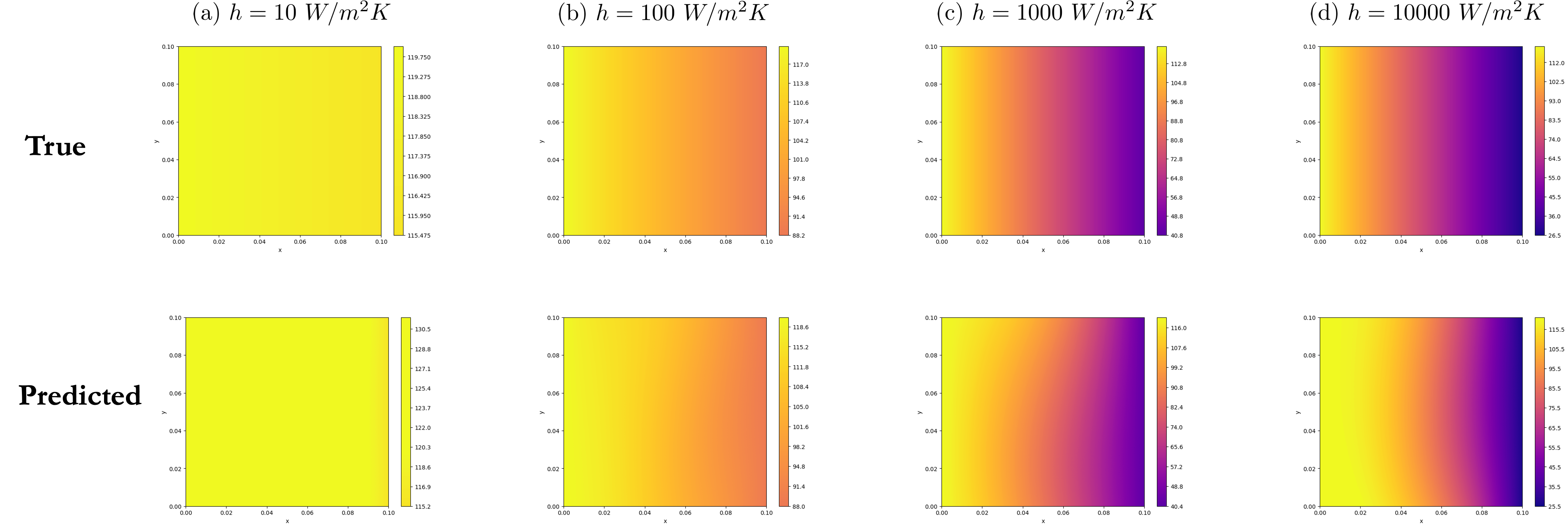}
    \caption{Plots showing the solution on a rectangular plate for  heat transfer coefficients (a) 10 $W/m^{2}K$, (b) 100 $W/m^{2}K$, (c)1000 $W/m^{2}K$ and (d) 10000 $W/m^{2}K$.}
    \label{fig:A1}
\end{figure}
\begin{table}[h!]
    \centering
    \begin{tabular}{|c|c|c|}
        \hline
        \textbf{h ($W/m^{2}K$) value predicted} & \textbf{true h ($W/m^{2}K$) value} & \textbf{error \%} \\
        \hline
        10.35 & 10 & 3.5 \\
        98.731 & 100 & 1.269 \\
        998.26 & 1000 & 0.174 \\
        9942.24 & 10000 & 0.5776 \\
        \hline
    \end{tabular}
    \caption{Comparison of Predicted and True h Values with Error Percentage}
    \label{tab:table1}
\end{table}

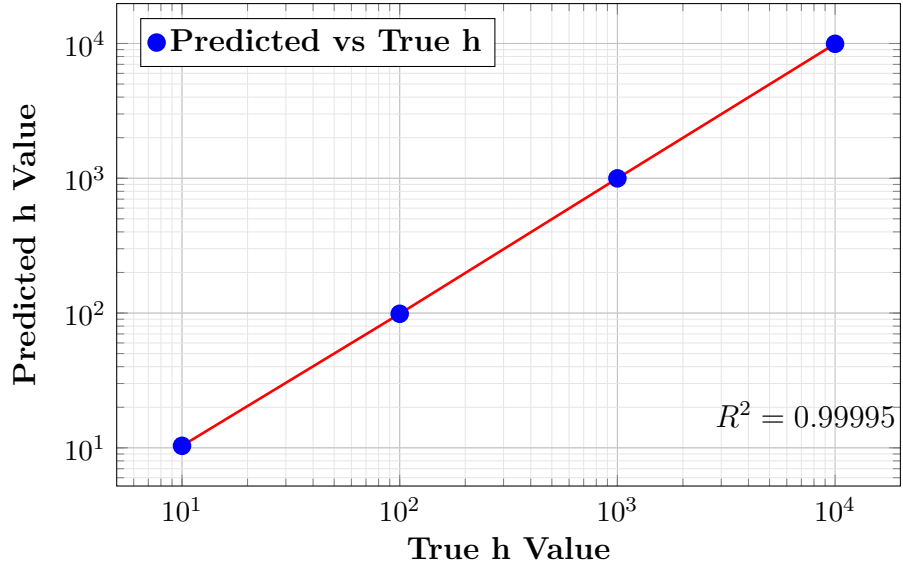
\begin{figure}[h!]
    \centering
    \begin{tikzpicture}
        \begin{axis}[
            width=12cm, height=8cm,
            xlabel={\textbf{True h Value}},
            ylabel={\textbf{Predicted h Value}},
            legend pos=north west,
            ymode=log,
            xmode=log,
            log basis y=10,
            log basis x=10,
            grid=both,
            minor tick num=1,
            grid style={line width=.1pt, draw=gray!20},
            major grid style={line width=.2pt, draw=gray!50},
            tick label style={font=\small},
            label style={font=\bfseries},
        ]
        % Plot the data points
        \addplot[
            color=blue,
            mark=*,
            mark size=3,
            line width=1pt,
            only marks,
        ] coordinates {
            (10, 10.35)
            (100, 98.731)
            (1000, 998.26)
            (10000, 9942.24)
        };

        % Connect the points with a line
        \addplot[
            color=red,
            line width=1pt
        ] coordinates {
            (10, 10.35)
            (100, 98.731)
            (1000, 998.26)
            (10000, 9942.24)
        };

        \addlegendentry{\textbf{Predicted vs True h}}
        
        % Display R^2 value
        \node[anchor=south west, font=\bfseries] at (rel axis cs:0.75,0.1) {$R^2 = 0.99995$};
    \end{axis}
    \end{tikzpicture}
    \caption{Log-Log Plot of True vs Predicted h Values with $R^2$ Value}
    \label{fig:plot1}
\end{figure}

Figure~\ref{fig:A1} shows the prediction of the temperature profile for different values of $h$. Table~\ref{tab:table1} and Figure~\ref{fig:plot1} show the comparison between predicted and true $h$ values. This example clearly shows the capabilities of PINNs to predict heat transfer coefficient $h$ for a range of values while predicting the correct temperature profile. 

A point to note here is that when imposing the loss function for this problem, imposing just the convective boundary condition as shown in Eqn.~(81) is not enough for uniqueness as there are   infinitely many pairs of $T(x,y)$ and $h$ that can satisfy the PDE and the convective boundary conditions. Therefore we additionally impose Eqn.~(84) in the loss function to guarantee uniqueness. This is important: we follow a similar principle for the experimental setup, where uniqueness is imposed by the conservation of energy principle (total heat in equals total heat out) along with the convective boundary condition at the inner surface of the pipes.

\subsubsection*{Part B: Experimental example}
%%% Add a description about the experiment and data and no data casses

\paragraph{Experimental set-up.}

\begin{figure}[btp]
\centering\includegraphics[width=0.6\textwidth]{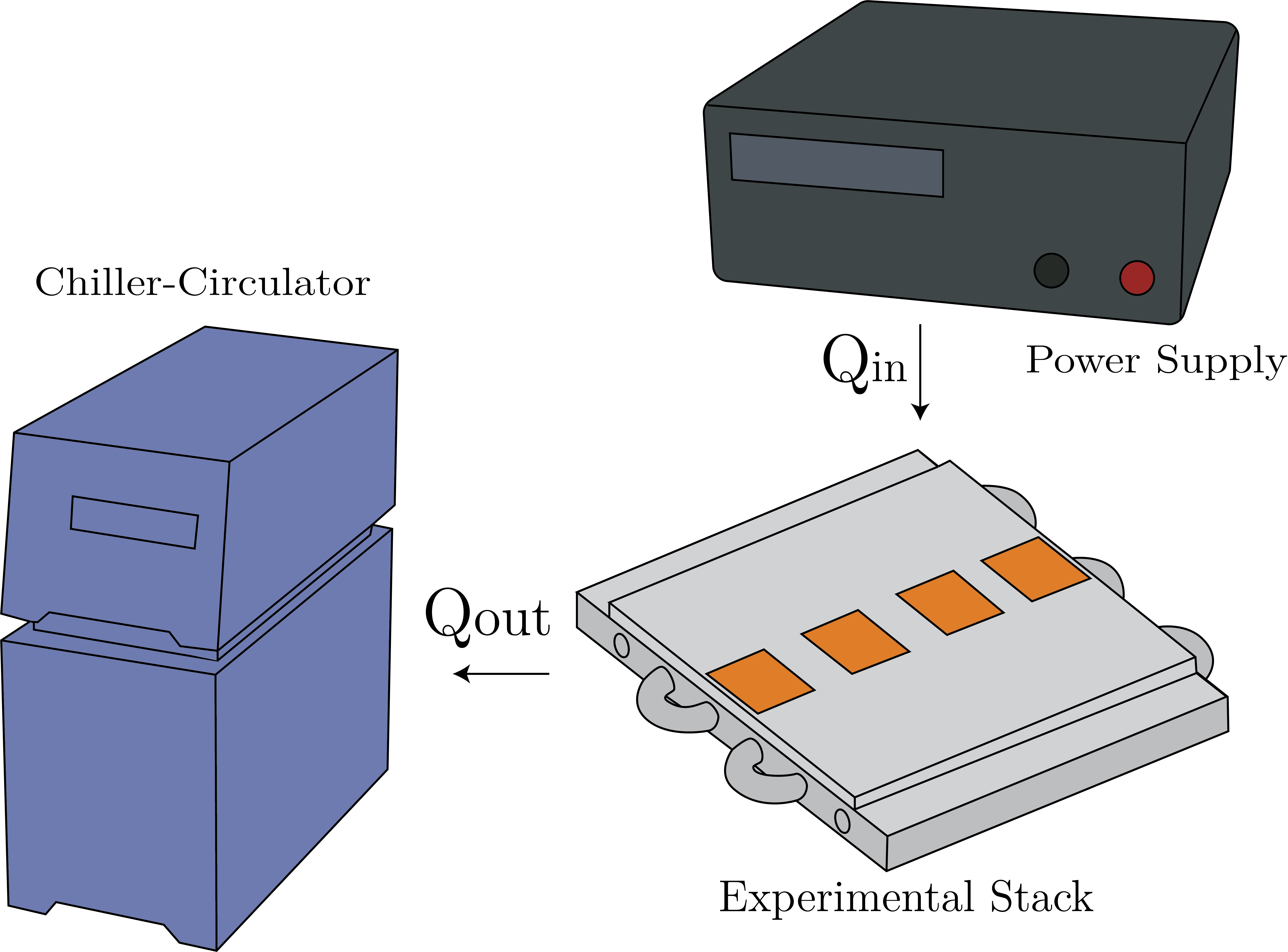} 
\caption{Graphic showing the experimental set up. The power resistors shown in orange on the experimental stack were heated by two Extech Adjustable Switching Mode Power Supplies. Water was circulated through the cold plate in the stack and the IKA RC 2 Basic Chiller-Circulator with controllable inlet temperature and flow rate. The heat flows in and out of the system are labeled as $Q_{in}$ and $Q_{out}$.}
\label{fig:exp}
\end{figure}

\begin{figure}[btp]
    \centering
    \includegraphics[width=0.5\linewidth]{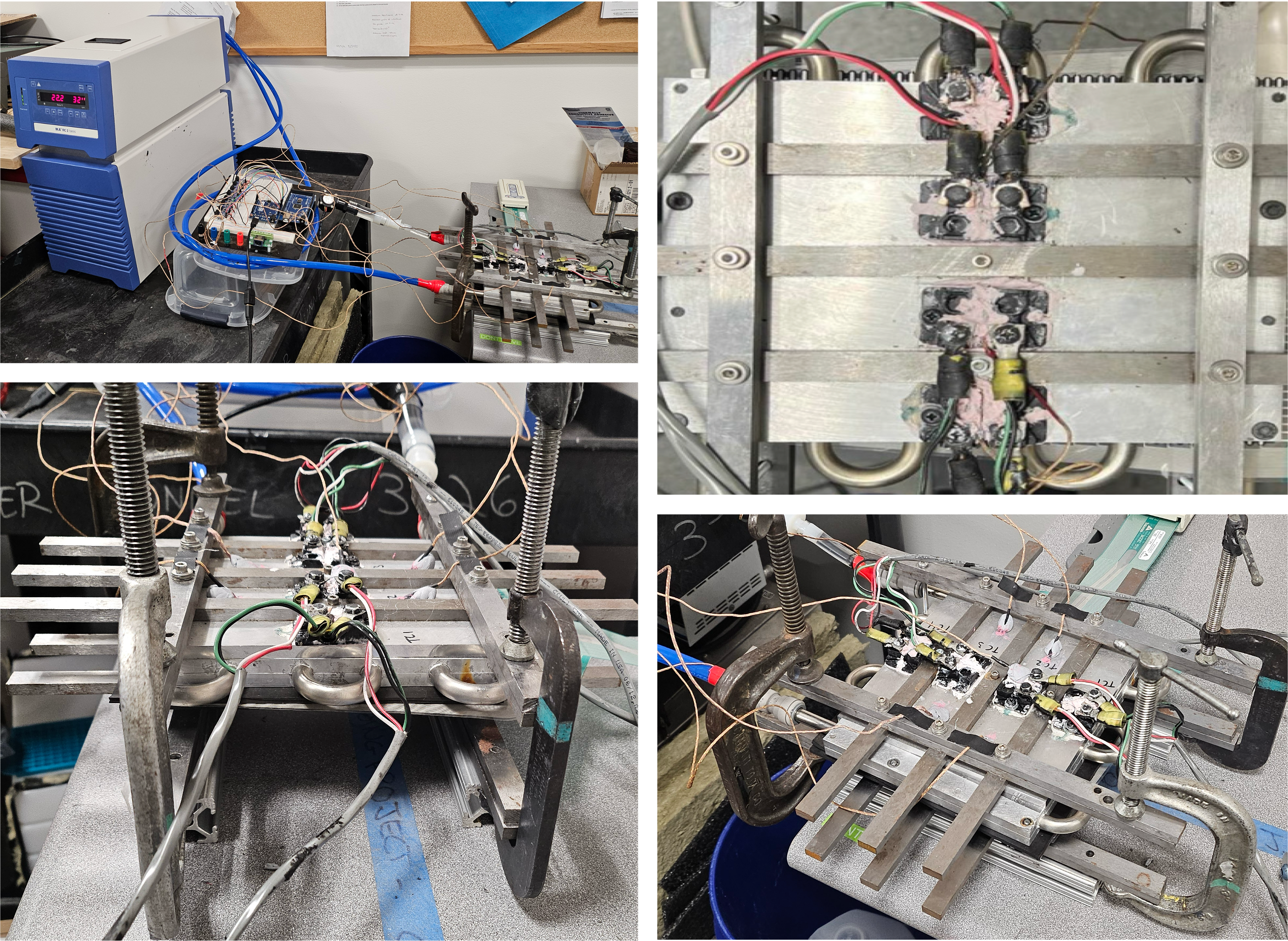}
    \caption{Photographs showing the experimental setup \cite{alvarez2024experimental}.}
    \label{fig:exp_setup}
\end{figure}

The physical experimental set-up is shown in Figures~\ref{fig:exp} and~\ref{fig:exp_setup}. $Q_{in}$ represents the heat flux delivered by the two power supplies to the power resistors in the experimental stack. $Q_{out}$ represents the heat flux removed by the chiller-circulator from the experimental stack through the cold plate. Water was used as the liquid in the chiller-circulator cold plate loop. 

Two Extech Adjustable Switching Mode Power Supplies were wired to two resistors each, providing up to 520 Watts to the experimental stack. The cold plate in the experimental stack was connected through tubing to an IKA RC 2 Basic Chiller-Circulator, which was used to control inlet temperature and flow rate. The Digiten flow sensor was placed in the outlet flow path to measure the volumetric flow rate of the system. 

To verify experimental repeatability, ten minute trials were taken three times over the course of several days for the lowest and highest power of the data set at three different flow rates. Once repeatability was established, the experimental conditions were expanded to a total of four different powers, five different flow rates, and three inlet temperatures. The wide variety of experimental conditions were selected to provide a robust training and test data set for the PINNs model.  Additional details regarding the experiments are available in \cite{alvarez2024experimental}.

\begin{figure}[hbtp]
    \centering
    \includegraphics[width=0.8\linewidth]{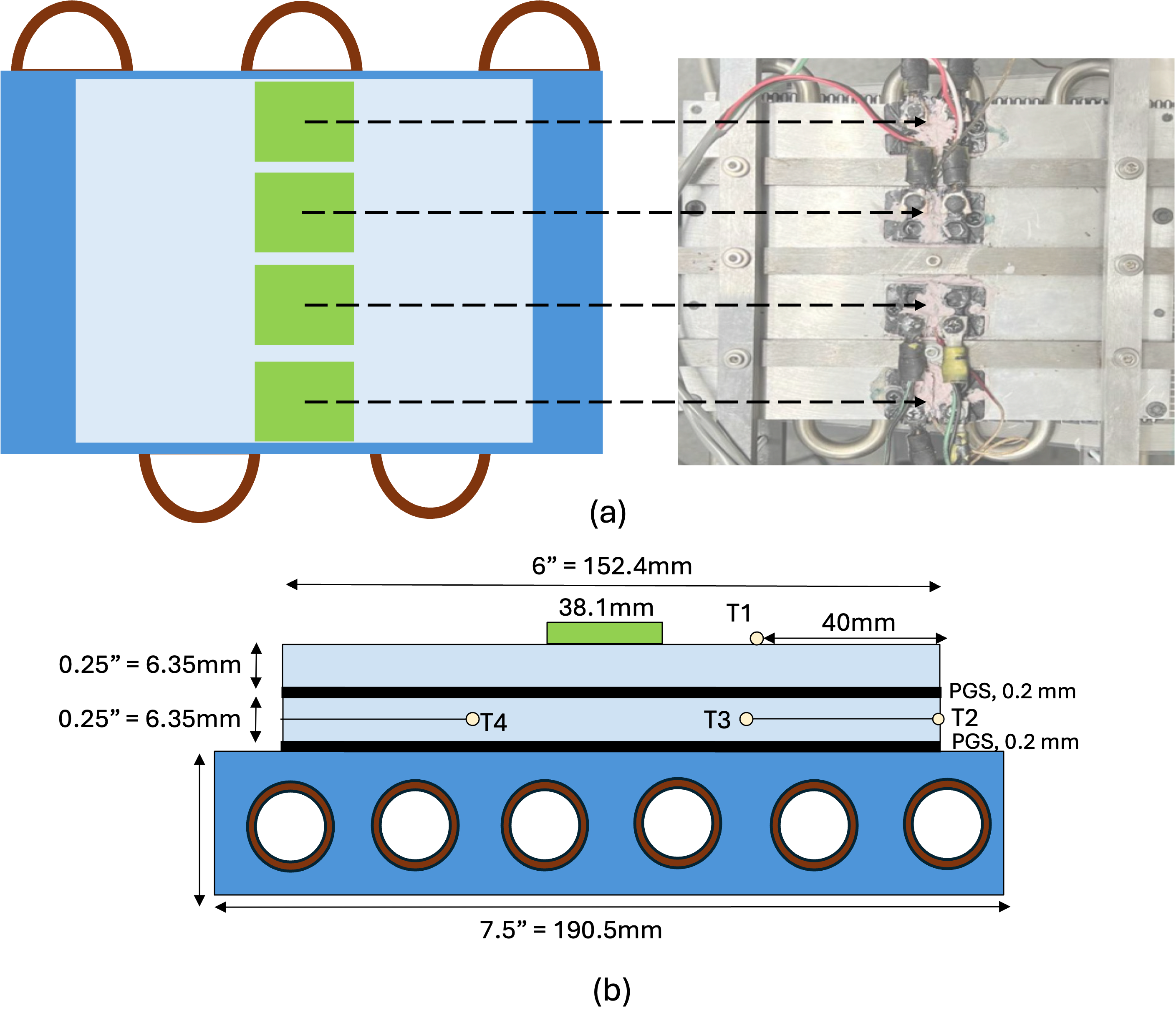}
    \caption{(a) A photograph (right) and a schematic diagram (left) of the experimental setup, plan view. (b) A schematic diagram of the cross-section used in PINNs simulation, showing  the measurement points for experimental data. T1, T2, T3, and T4 correspond to the Face, Side, In2 and In1 data points in the experiment, respectively.}
    \label{fig:exp_data}
\end{figure}

Figure~\ref{fig:exp_data} shows the sampling location of temperature data during the experiments; these temperatures are reported in the final row of Tables 2 through 5 in the main text and Table 7 through 9 in the Appendix.

\paragraph{Experimental and Numerical Results.}

To validate and test the PINNs results, we consider 9 different experimental cases, in which we compare the computed velocity from the PINNs approach to the actual velocity of the experimental values. The details of the various experimental runs are shown in Table 6 in Appendix A. The variations are for different power values as well as for different inlet and outlet temperature values. The power values vary from 151.8 Watts to 259.2 Watts, the inlet temperature varies from 10.0226 $^{\circ}C$ to 14.9866 $^{\circ}C$ and the outlet temperature varies from  12.55 $^{\circ}C$ to 21.20 $^{\circ}C$. For all the experiments, the number of points for training the PINNs setup was as follows : 7624 residual points for the cold plate, 6000 residual points for all the other layers except the top layer where 7000 residual points were used for training. Moreover 200  points were used for both the right and left hand side boundaries of all the layers. For the bottom boundary of the cold plate and 200 points were used and for the top boundary, with the incoming heat flux, 300 points were used. 200 points were used for all the interfaces across layers. All these points were selected using the Latin hyper cube sampling method \cite{shields2016generalization}.

To ensure robustness and reliability of the predicted results, three simulation runs were conducted with different seeds for each case. This is done in order to account for the variability due to model initialization and fluctuations during the optimization process. Taking an average across three trials reduces the effect of potential outliers and leads to a more stable prediction of the heat transfer coefficients, velocity fields and temperature distributions.

\begin{figure}[hbtp]
    \centering
    \includegraphics[width=0.7\textwidth]{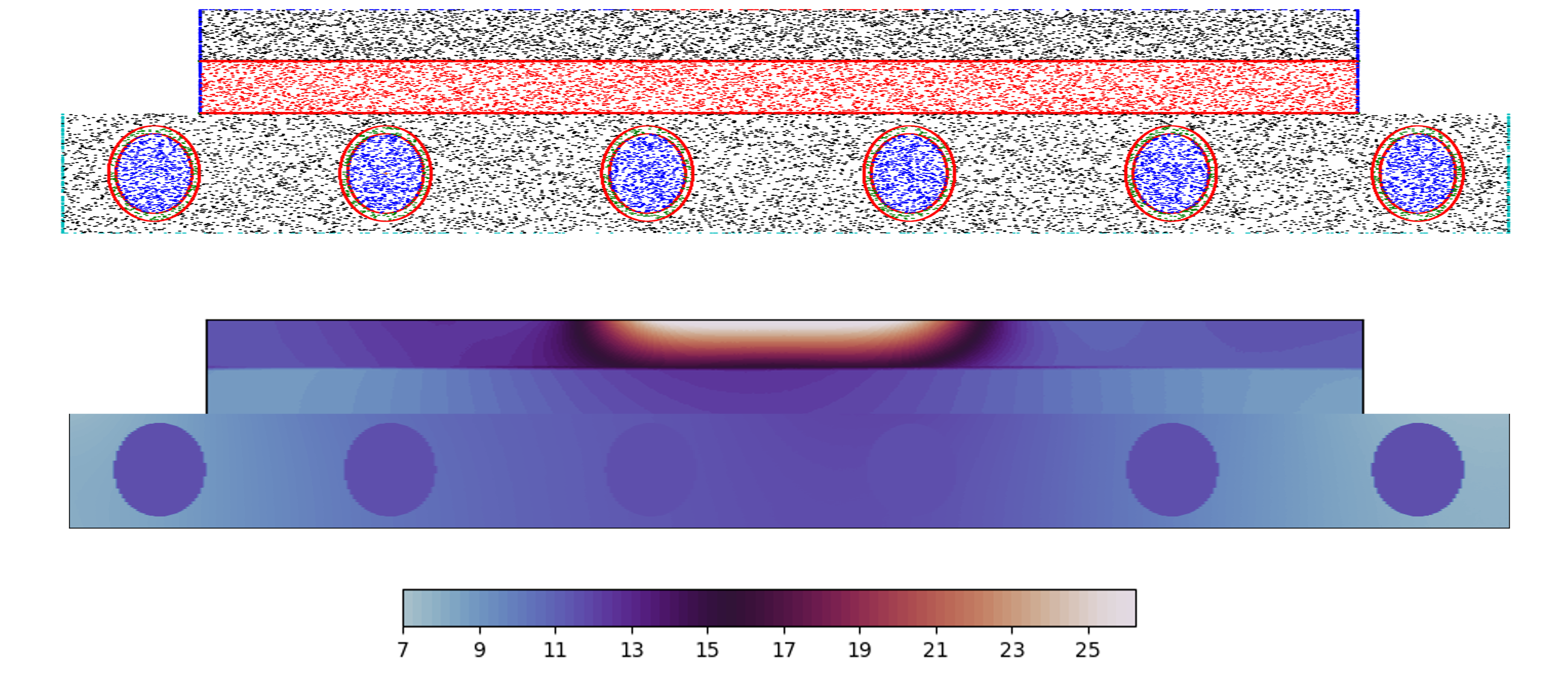}
    \caption{Figure showing the training points used for PINNs and the temperature distribution for case A13\_4 when no data is used for training. The temperature shown here are in $^{\circ}C$.}
    \label{fig:13_4_no_data}
\end{figure}

\begin{table}[hbtp]
\centering
\caption{Case A13\_4 \hspace{0.2cm} (No data used, $r = 0.005$ mm, $C_p = 4188.5 \, J/Kg-K$, $\rho = 999.1 \, Kg/m^3$)}
\label{tab:case13_4}
\resizebox{\textwidth}{!}{%
\begin{tabular}{cccccccccccc}
\toprule
\multirow{2}{*}{\textbf{Trial}} & \multirow{2}{*}{\textbf{$h_{nn}$ (W/m$^2$K)}} & \multirow{2}{*}{\textbf{$v_{nn}$ (m/s)}} & \multirow{2}{*}{\textbf{$v_{exp}$ (m/s)}} & \multicolumn{2}{c}{\textbf{Side ($^\circ$C)}} & \multicolumn{2}{c}{\textbf{Face ($^\circ$C)}} & \multicolumn{2}{c}{\textbf{In1 ($^\circ$C)}} & \multicolumn{2}{c}{\textbf{In2 ($^\circ$C)}} \\
& & & & \textbf{Pred} & \textbf{Exp} & \textbf{Pred} & \textbf{Exp} & \textbf{Pred} & \textbf{Exp} & \textbf{Pred} & \textbf{Exp} \\
\midrule
1 & 3170.89  & 0.33  & 0.296 & 13.52  & 23.57  & 14.65  & 27.48  & 13.56  & 25.85  & 13.89  & 25.90 \\
2 & 3281.05  & 0.32  & 0.296 & 13.65  & 23.57  & 13.91  & 27.48  & 13.10  & 25.85  & 13.35  & 25.90 \\
3 & 3165.11  & 0.34  & 0.296 & 14.65  & 23.57  & 14.21  & 27.48  & 14.23  & 25.85  & 13.47  & 25.90 \\
\midrule
\textbf{Mean} & 3205.68  & 0.33  & 0.296 & 13.94  & 23.57  & 14.25  & 27.48  & 13.63  & 25.85  & 13.57  & 25.90 \\
\textbf{Std}  & 65.34    & 0.01  &       & 0.62   & -      & 0.37   & -      & 0.56   & -      & 0.28   & -      \\
\bottomrule
\end{tabular}%
}
\end{table}

%\begin{table}[hbp]
%\centering
%\caption{Case A13\_4 \hspace{0.2cm} (No data used, $Q_{in} = 259.2$ W, $r = 0.005$ mm, $C_p = 4188.5 \, J/Kg-K$, $\rho = 999.1 \, Kg/m^3$)}
%\label{tab:case13_4}
%\resizebox{\textwidth}{!}{%
%\begin{tabular}{ccccccccccc}
%\toprule
%\textbf{Trial no.} & \textbf{$h_{nn}$ (W/m2K)} & \textbf{$v_{nn}$ (m/s)} & \textbf{$v_{exp}$ (m/s)} & \textbf{Side} & \textbf{Face} & \textbf{In1} & \textbf{In2}  & \textbf{$Q_{in}$} & \textbf{$Q_{out}$} \\
%\midrule
%1 & 3170.89 & 0.33114 & 0.296 & 13.52 & 14.65 & 13.56  & 13.89   & 259.2 & 230.53  \\
%2 & 3281.05 & 0.322 & 0.296 & 13.65 & 13.91 & 13.1 &  13.35 & 259.2 &  238.54\\
%3 & 3165.11 & 0.34577 & 0.296 & 14.65 & 14.21 &  14.23  & 13.47 & 259.2 & 230.11 \\
%\midrule
%\textbf{mean} & 3205.683333 & 0.33297 & 0.296 & 13.94 & 14.25 & 13.63 & 13.57  & 259.2 & 233.06 \\
%\textbf{std} & 65.3333983 & 0.0119902 &  & 0.618304132 & 0.37 & 0.56   & 0.28 & 0 &  4.75\\
%\midrule
%\multicolumn{11}{l}{\textbf{Experimental Data}} \\
%\midrule
%\textbf{Side} & 23.571 & \multicolumn{2}{c}{\textbf{Face}} & 27.4784 & \multicolumn{2}{c}{\textbf{In1}} & 25.8598 & \multicolumn{2}{c}{\textbf{In2}} & 25.9001 \\
%\bottomrule
%\end{tabular}%
%}
%\end{table}

The first example is Case A13\_4, in which the framework was run without any specific input data.  Results are shown in Figure~\ref{fig:13_4_no_data} and Table~\ref{tab:expdata1}. 
The mean velocity predicted in this case (0.33 m/s) is close to the experimental velocity (0.29 m/s), but significant differences are observed in the temperature values along the side, face, in1, and in2. %The temperature profile for this case is shown in Figure~\ref{fig:13_4_no_data}. 
This is because the velocity loss is mainly governed by the $L_Q$ as shown in Eqn (44). Consequently, accurate velocity predictions can be achieved without data, but at the expense of precise local temperature information.

\begin{figure}[hbtp]
    \centering
    \includegraphics[width=0.7\textwidth]{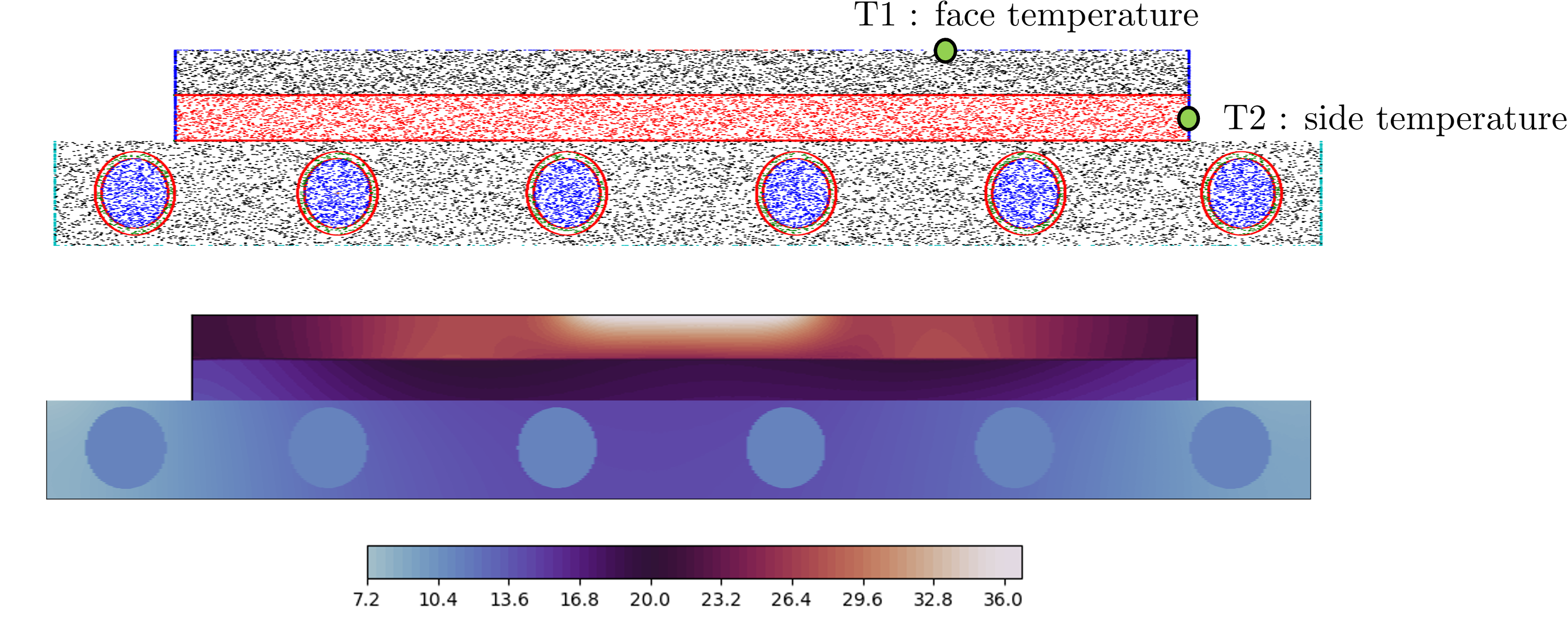}
    \caption{Figure showing the training points used for PINNs and the temperature distribution for case A13\_4 when data from face (T1) and sides (T2) are used for training. The mean velocity predicted in this case (~0.30 m/s) is close to the experimental velocity (~0.29 m/s) with close match with the temperature values along In1 and In2. For detailed information look at table 3. The temperature shown here are in $^{\circ}C$.}
    \label{fig:13_4_data}
\end{figure}

%%%%%%%%%%%%%%%%%%%%%%%%%%% new version of table %%%%%%%%%%%%%%%%%%%%%%%%%
\begin{table}[hbtp]
\centering
\caption{Case A13\_4 \hspace{0.2cm} (Face and Side temp data used, $r = 0.005$ mm, $C_p = 4188.5 \, J/Kg-K$, $\rho = 999.1 \, Kg/m^3$)}
\label{tab:expdata1}
\resizebox{\textwidth}{!}{%
\begin{tabular}{cccccccccccc}
\toprule
\multirow{2}{*}{\textbf{Trial}} & \multirow{2}{*}{\textbf{$h_{nn}$ (W/m$^2$K)}} & \multirow{2}{*}{\textbf{$v_{nn}$ (m/s)}} & \multirow{2}{*}{\textbf{$v_{exp}$ (m/s)}} & \multicolumn{2}{c}{\textbf{Side ($^\circ$C)}} & \multicolumn{2}{c}{\textbf{Face ($^\circ$C)}} & \multicolumn{2}{c}{\textbf{In1 ($^\circ$C)}} & \multicolumn{2}{c}{\textbf{In2 ($^\circ$C)}} \\
& & & & \textbf{Pred} & \textbf{Exp} & \textbf{Pred} & \textbf{Exp} & \textbf{Pred} & \textbf{Exp} & \textbf{Pred} & \textbf{Exp} \\
\midrule
1 & 2913.79 & 0.28 & 0.296 & 22.41  & 23.57  & 26.77  & 27.47  & 23.01  & 25.86  & 23.54  & 25.90 \\
2 & 2957.04 & 0.29 & 0.296 & 22.35  & 23.57  & 27.11  & 27.47  & 22.81  & 25.86  & 22.99  & 25.90 \\
3 & 2918.29 & 0.28 & 0.296 & 23.11  & 23.57  & 26.42  & 27.47  & 24.32  & 25.86  & 24.51  & 25.90 \\
\midrule
\textbf{Mean} & 2929.71 & 0.28 & 0.296 & 22.62  & 23.57  & 26.76  & 27.47  & 23.38  & 25.86  & 23.68  & 25.90 \\
\textbf{Std}  & 23.77   & 0.005 & -  & 0.422   & -       & 0.345   & -       & 0.82    & -       & 0.78    & -       \\
\bottomrule
\end{tabular}%
}
\end{table}

%%%%%%%%%%%%%%%%%%%%%%%%%%%%%%%%%%%%%%%%%%%%%%%%%%%%%%%%%%%%%%%%%%%%%%%%%%

\begin{figure}[hbtp]
    \centering
    \includegraphics[width=0.5\linewidth]{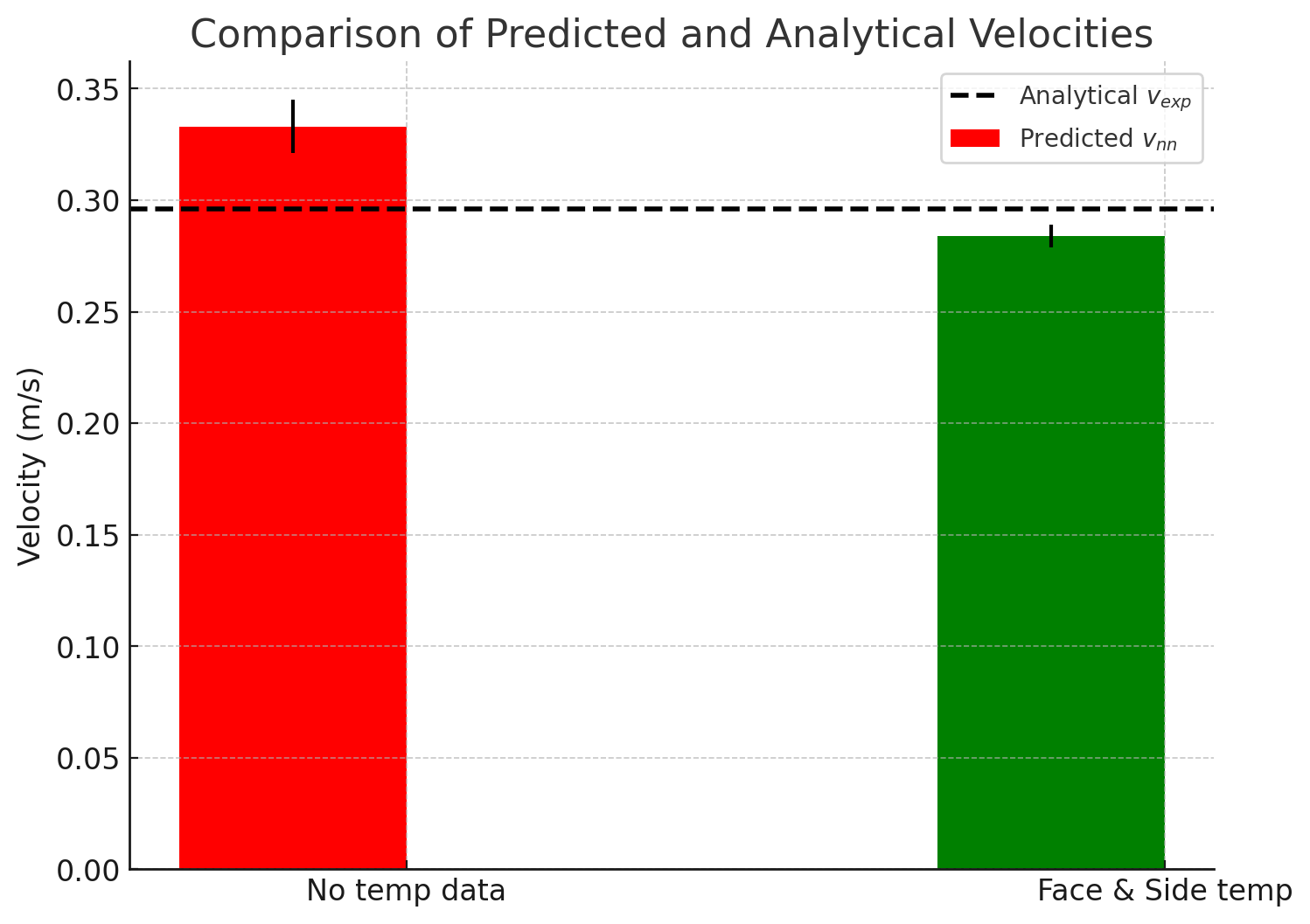}
    \caption{Comparison of velocity for the Case A13\_4 with and without temperature constraint.}
    \label{fig:v_compare}
\end{figure}

\begin{figure}[hbtp]
    \centering
    \includegraphics[width=0.5\linewidth]{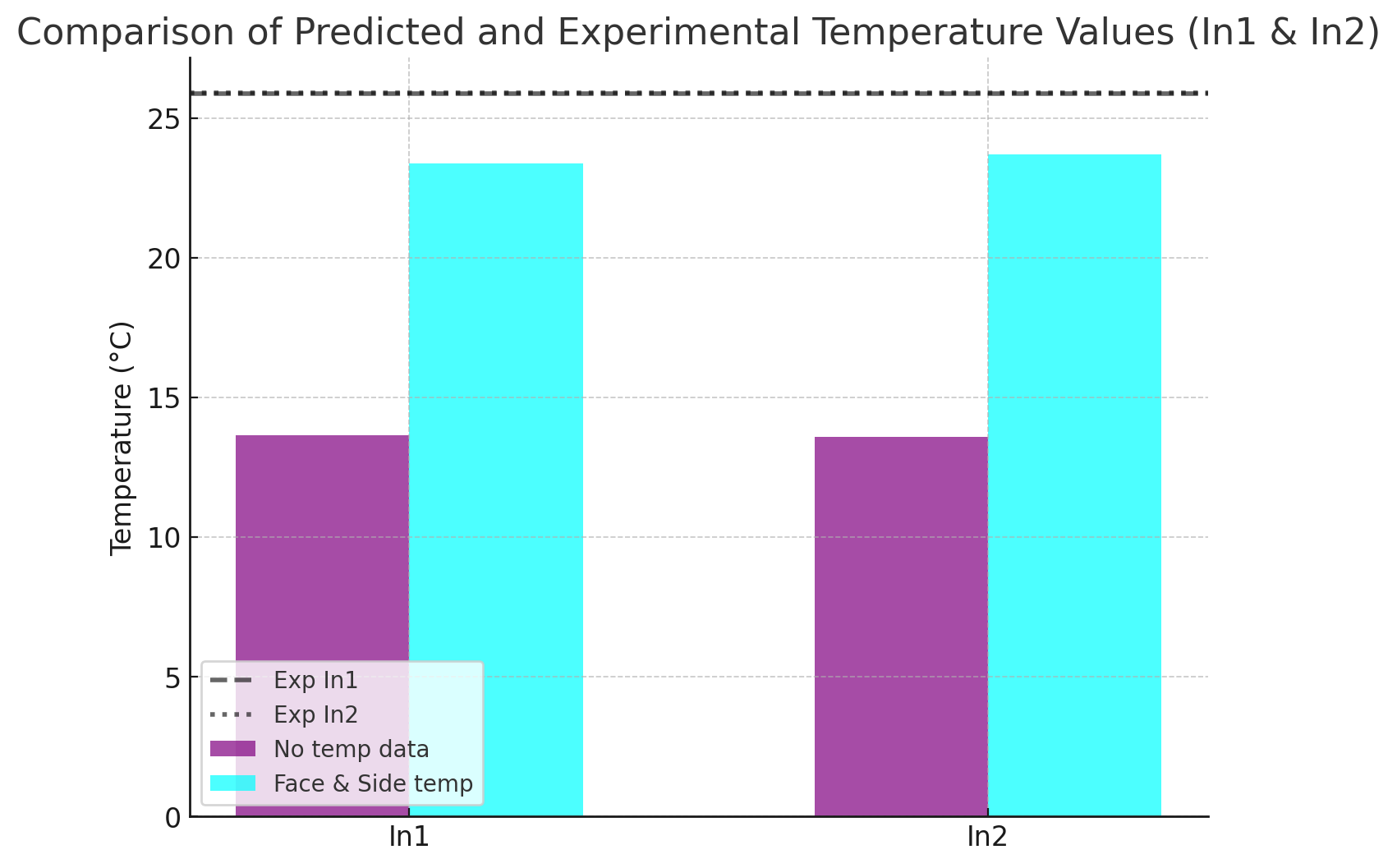}
    \caption{Comparison of temperature predicted at in1 and in2 for the Case A13\_4 with and without temperature constraint.}
    \label{fig:ln_comp}
\end{figure}

Next we run the same setup but this time we include temperature measurements from two sensors (T1 and T2 in Fig 15 (b)) that provide information about the face temperature and side temperature. Results are shown in Figure~\ref{fig:13_4_data} and Table~\ref{tab:expdata1}.  The results indicate that when temperature data were not included, the model overestimated the heat transfer coefficient $h_{nn}$ (mean value of 3205.68 W/m²K) compared to the temperature-constrained case (2929.71 W/m²K). This suggests that the addition of temperature constraints moderates the predicted thermal resistance, leading to a more physically consistent solution. Similarly, the predicted velocity $v_{nn}$ was higher than the experimental value (0.33297 m/s) when temperature data were absent, whereas including temperature data reduced this discrepancy, bringing the predicted velocity closer to the experimental measurement of 0.296 m/s (see Fig~\ref{fig:v_compare}). Also, we can see a significant improvement in the prediction of temperature values  
in1 and in2 at locations T3 and T4 respectively.  The temperatures were significantly underestimated in the 
no-data case compared to the experimental data, whereas in the temperature-informed case, the predicted values closely matched the experimental measurements (see Fig~\ref{fig:ln_comp}). This indicates that the inclusion of thermal boundary data improves the fidelity of temperature predictions throughout the entire domain.

\begin{figure}[!ht]
    \centering
    \includegraphics[width=0.7\textwidth]{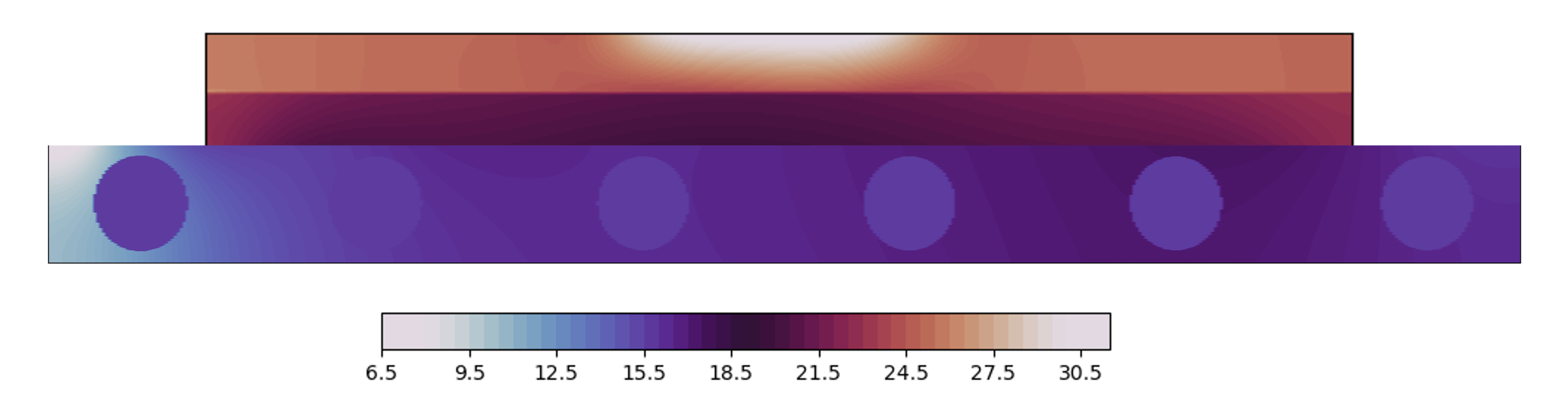}
    \caption{Figure showing the temperature distribution for case A11\_1 when data from face and sides are used for training. The mean velocity predicted in this case (0.24~m/s) is close to the experimental velocity (0.29~m/s) with close match with the temperature values along In1 and In2. For detailed information look at table 4. The temperature shown here are in $^{\circ}C$.}
    \label{fig:11_1_data}
\end{figure}

\begin{figure}[!ht]
    \centering
    \includegraphics[width=0.7\textwidth]{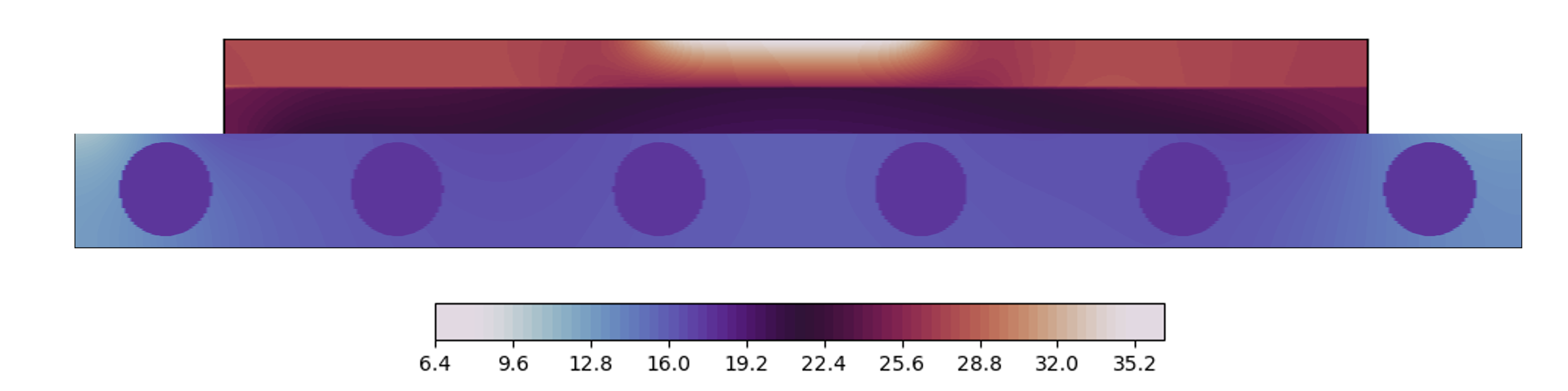}
    \caption{Figure showing the temperature distribution for case A13\_3 when data from face and sides are used for training. The mean velocity predicted in this case (~0.64 m/s) is close to the experimental velocity (~0.69 m/s) with close match with the temperature values along In1 and In2. For detailed information look at table 5. The temperature shown here are in $^{\circ}C$.}
    \label{fig:13_3_data}
\end{figure}

Additional experimental configurations were simulated, the results of which can be found below (Case A11\_1 and Case A13\_3) and in the Appendix section (Case A12\_2, A14\_2, A13\_7). The 
temperature profiles for Case A11\_1 and Case A13\_3 are shown in Figure~\ref{fig:11_1_data} and~\ref{fig:13_3_data} respectively. In each of these cases, one of the parameters such as $Q_{in}$, $T_{in}$ and $T_{out}$ was varied to assess the robustness of the velocity predicted by the proposed method.  Across all the scenarios, the predicted velocity matches closely with the experimental velocity.

\begin{table}[h!]
\centering
\caption{Case A11\_1 \hspace{0.2cm} (Face and Side temp data used, $Q_{in} = 151.80$ W, $r = 0.005$ mm, $C_p = 4188.5 \, J/Kg-K$, $\rho = 999.1 \, Kg/m^3$)}
\label{tab:expdata3}
\resizebox{\textwidth}{!}{%
\begin{tabular}{cccccccccccc}
\toprule
\multirow{2}{*}{\textbf{Trial}} & \multirow{2}{*}{\textbf{$h_{nn}$ (W/m$^2$K)}} & \multirow{2}{*}{\textbf{$v_{nn}$ (m/s)}} & \multirow{2}{*}{\textbf{$v_{exp}$ (m/s)}} & \multicolumn{2}{c}{\textbf{Side ($^\circ$C)}} & \multicolumn{2}{c}{\textbf{Face ($^\circ$C)}} & \multicolumn{2}{c}{\textbf{In1 ($^\circ$C)}} & \multicolumn{2}{c}{\textbf{In2 ($^\circ$C)}} \\
& & & & \textbf{Pred} & \textbf{Exp} & \textbf{Pred} & \textbf{Exp} & \textbf{Pred} & \textbf{Exp} & \textbf{Pred} & \textbf{Exp} \\
\midrule
1 & 910.49  & 0.324  & 0.296 & 22.88  & 22.87  & 25.31  & 25.27  & 23.01  & 24.00  & 21.44  & 23.98 \\
2 & 1066.23 & 0.304  & 0.296 & 21.91  & 22.87  & 25.61  & 25.27  & 22.81  & 24.00  & 23.85  & 23.98 \\
3 & 1085.77 & 0.323  & 0.296 & 22.88  & 22.87  & 25.28  & 25.27  & 24.32  & 24.00  & 21.22  & 23.98 \\
\midrule
\textbf{Mean} & 1020.60 & 0.317  & 0.296 & 22.62  & 22.87  & 25.40  & 25.27  & 23.38  & 24.00  & 22.17  & 23.98 \\
\textbf{Std}  & 95.81   & 0.011  & -       & 0.0085 & -      & 0.182  & -      & 0.82   & -      & 1.459  & -      \\
\bottomrule
\end{tabular}%
}
\end{table}

\begin{table}[h!]
\centering
\caption{Case A13\_3 \hspace{0.2cm} (Using Face and Side Data, $Q_{in} = 259.2$ W, $r = 0.005$ mm, $C_p = 4188.5 \, J/Kg-K$, $\rho = 999.1 \, Kg/m^3$)}
\label{tab:expdata4}
\resizebox{\textwidth}{!}{
\begin{tabular}{ccccccccccc}
\toprule
\multirow{2}{*}{\textbf{Trial}} & \multirow{2}{*}{\textbf{$h_{nn}$ (W/m$^2$K)}} & \multirow{2}{*}{\textbf{$v_{nn}$ (m/s)}} & \multicolumn{2}{c}{\textbf{Side ($^\circ$C)}} & \multicolumn{2}{c}{\textbf{Face ($^\circ$C)}} & \multicolumn{2}{c}{\textbf{In1 ($^\circ$C)}} & \multicolumn{2}{c}{\textbf{In2 ($^\circ$C)}}  \\
& & & \textbf{Pred} & \textbf{Exp} & \textbf{Pred} & \textbf{Exp} & \textbf{Pred} & \textbf{Exp} & \textbf{Pred} & \textbf{Exp}  \\
\midrule
1 & 1618.65  & 0.71  & 25.19  & 25.19  & 29.13  & 29.12  & 26.95  & 27.25  & 26.86  & 27.23   \\
2 & 1542.34  & 0.73  & 25.16  & 25.19  & 29.09  & 29.12  & 25.54  & 27.25  & 25.74  & 27.23   \\
3 & 1600.91  & 0.68  & 25.19  & 25.19  & 29.12  & 29.12  & 26.22  & 27.25  & 26.31  & 27.23   \\
\midrule
\textbf{Mean} & 1587.30  & 0.71  & 25.18  & 25.19  & 29.12  & 29.12  & 26.23  & 27.25  & 26.30  & 27.23   \\
\textbf{Std}  & 39.93    & 0.025 & 0.02   & -      & 0.022  & -      & 0.71   & -      & 0.56   & -       \\
\bottomrule
\end{tabular}%
}
\end{table}

%\paragraph{Comparison of velocity predicted with experimental result}
%\begin{figure}[!ht]
%    \centering
%    \includegraphics[width=\textwidth]{Figures/exp.png}
%    \caption{Comparison between PINNs prediction and experimental results}
%    \label{fig:enter-label}
%\end{figure}
\clearpage
\newpage

\section{Conclusions}
\label{conclusions}
We have presented a new framework and training strategy for using PINNs to determining optimal velocity for indirect liquid cooling, given the required inlet and outlet temperature of the cooling fluid. Rather than solving the full domain with the coupled Navier-Stokes equations, we solve the steady state heat equations in a cross-sectional area of the domain and use the heat transfer coefficient as an inverse parameter under forced convection boundary conditions to determine the velocity. The results of the proposed methodology align well with experimental results. This gives a simpler and faster way to determine the optimal velocity for such a setup. Moreover the same setup could be used to impose additional constraints such as temperature range for optimal operation, which cannot be done using traditional methods. Convergence of the proposed method to the analytical solution has been theoretically analyzed. Further research will focus on integrating the method with operator learning so that retraining for newer domains can be avoided.

\section*{Acknowledgments}
The authors greatfully acknowledge valuable discussions with Professor George Em Karniadakis of Brown University. I. K. Alvarez, J. Chalfant and C. Chryssostomidis were supported by the Office of Naval Research Grant No. N00014-21-1-2124, and by the National Oceanic and Atmospheric Administration (NOAA) Grant No. NA22OAR4170126.  Approved for public release under DCN\#2025-10-24-1624.

\bibliography{reference}
\bibliographystyle{abbrvnat}

\newpage

\appendix
\section{Appendix: Background about Fully Connected Neural Networks and Automatic Differentiation}
\subsection{Fully-connected Neural Networks (FNN)}
\begin{figure}[!ht] %% figure
\begin{center}
\includegraphics[width=.7\linewidth]{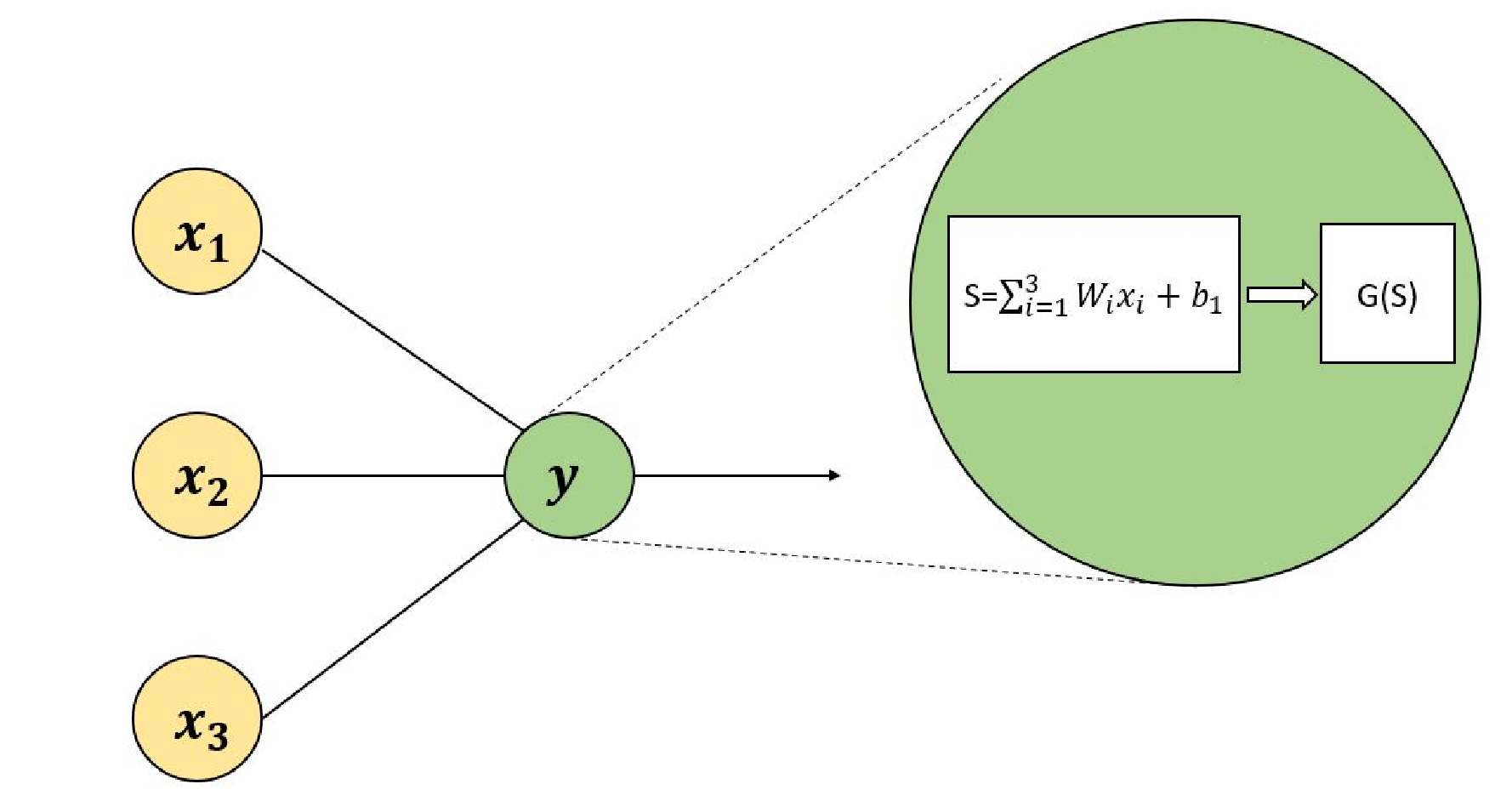}
\vspace{0.1in}
\caption{General operation of a perceptron.}
\label{Fig_4}
\end{center}
\end{figure}

Perceptrons are the fundamental building blocks of an FNN. It can be seen from Figure~\ref{Fig_4} that the perceptron performs two functions. One is to  multiply each input by weights, sum them up and then add a constant (bias) to them. The other is to pass the previously calculated value through an activation function $G$ which becomes the output for the given perceptron and an input for the next perceptron.  The process continues in this manner until the output layer is reached.

\begin{figure}[!ht] %% figure

\ \\
\vspace*{-.18in}

\begin{center}
\includegraphics[width=.9\linewidth]{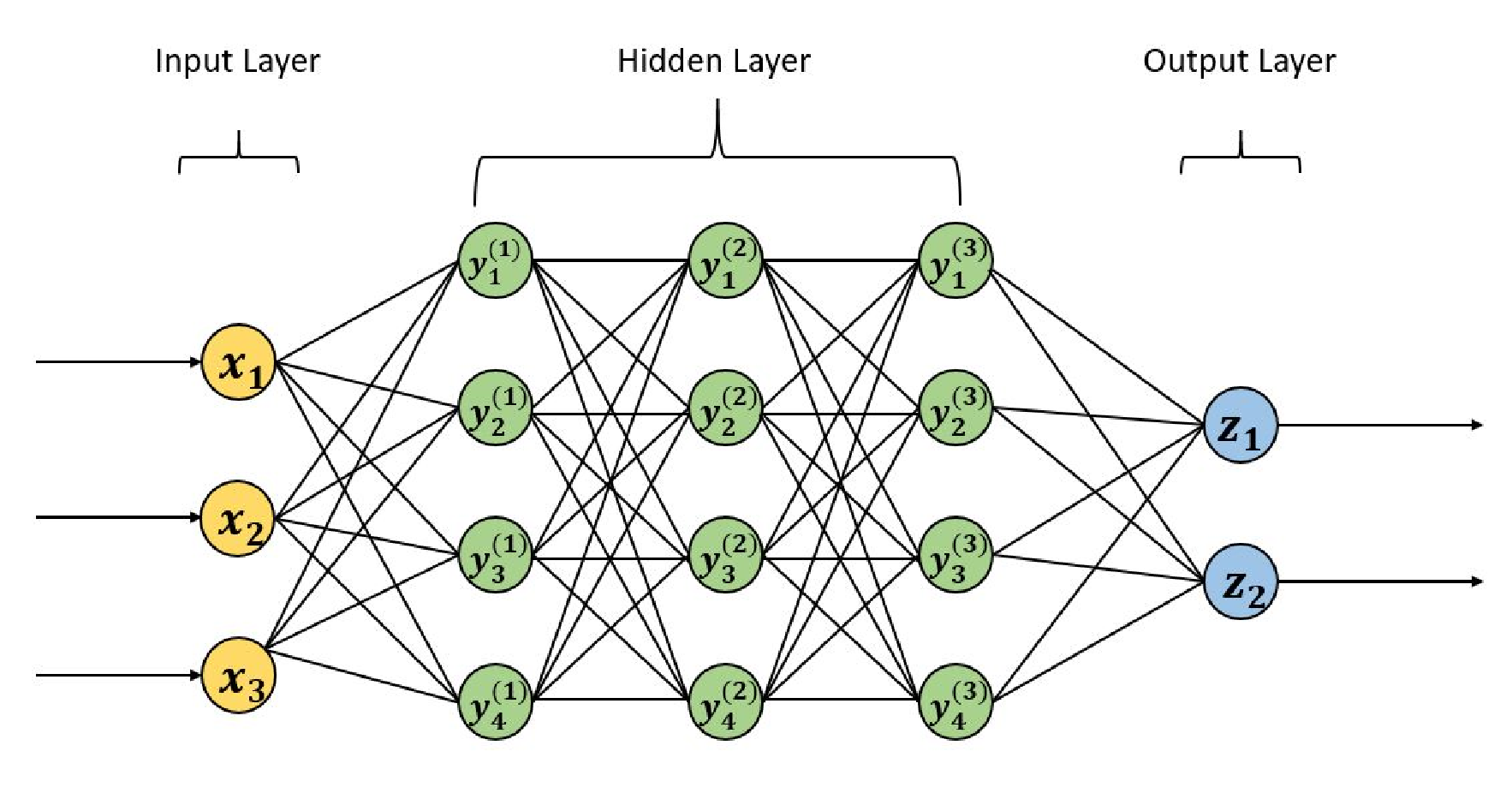}
\caption{General structure of a fully connected neural network.}
\label{Fig_3}
\end{center}
\end{figure}

As such, for the neural net in Figure~\ref{Fig_3}, we have
\begin{align}
    y_{i}^{(1)} &= G[\sum_{i=1}^{3}(W^{(1)}_{i}x_{i}+b^{(1)}_{i})],   \\
    y_{i}^{(2)} &= G[\sum_{i=1}^{4}(W^{(2)}_{i}y_{i}^{(1)}+b^{(2)}_{i})],   \\
    y_{i}^{(3)} &= G[\sum_{i=1}^{4}(W^{(3)}_{i}y_{i}^{(2)}+b^{(3)}_{i})],  
\end{align}and
\begin{align}
    z_{1} &= \sum_{i=1}^{4}W_{i}y_{i}^{(3)}+b_{1},  \\
    z_{2} &= \sum_{i=1}^{4}\Tilde{W}_{i}y_{i}^{(3)}+b_{2}, 
\end{align}where $W^{(j)}_{i}$, $b^{(j)}_{i}$, $W_{i}$, $\Tilde{W}_{i}$, $b_{1}$ and $b_{2}$ are the weights and biases involved in the neural network for $i,j=1,2,3$. 
\subsubsection*{Automatic Differentiation} 

Automatic differentiation (Autodiff)  is an elegant approach that can be used to calculate the partial derivatives of any arbitrary function in a given point. It decomposes the function in a sequence of elementary arithmetic operations (+, -, *, /) and functions (max, exp, log, cos, sin…); then uses the chain rule to work out the function’s derivative with respect to its initial parameters.

There are two variants of Autodiff:
\begin{itemize}
\item Forward-Mode, which is a hybrid of symbolic and numerical differentiation; while numerically precise, it requires one pass through the computational graph for each input parameter, which is resource consuming.
\item Reverse-Mode, on the other hand, only requires 2 passes through the computational graph to evaluate both the function and its partial derivatives.
\end{itemize}

\subsection{Reverse-Mode Autodiff}
\begin{figure}[tb]
    \centering
    \includegraphics[width=0.7\textwidth]{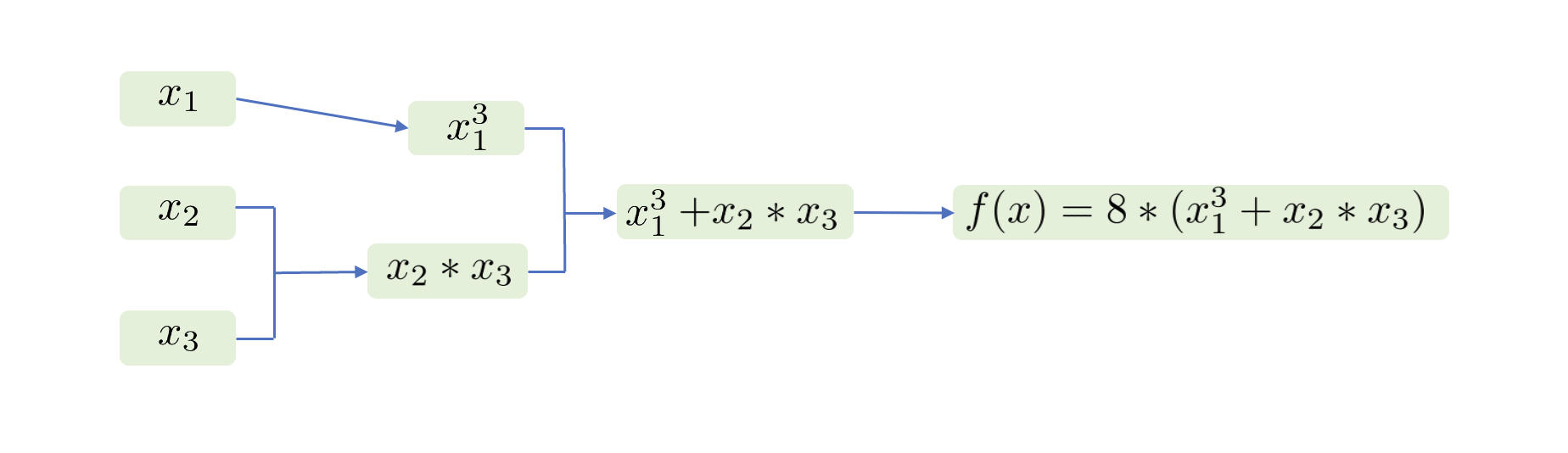}
    \caption{Computational graph of f(x).}
    \label{fig:CompGraph}
\end{figure}

\begin{figure}[bt]
    \centering
    \includegraphics[width=0.7\textwidth]{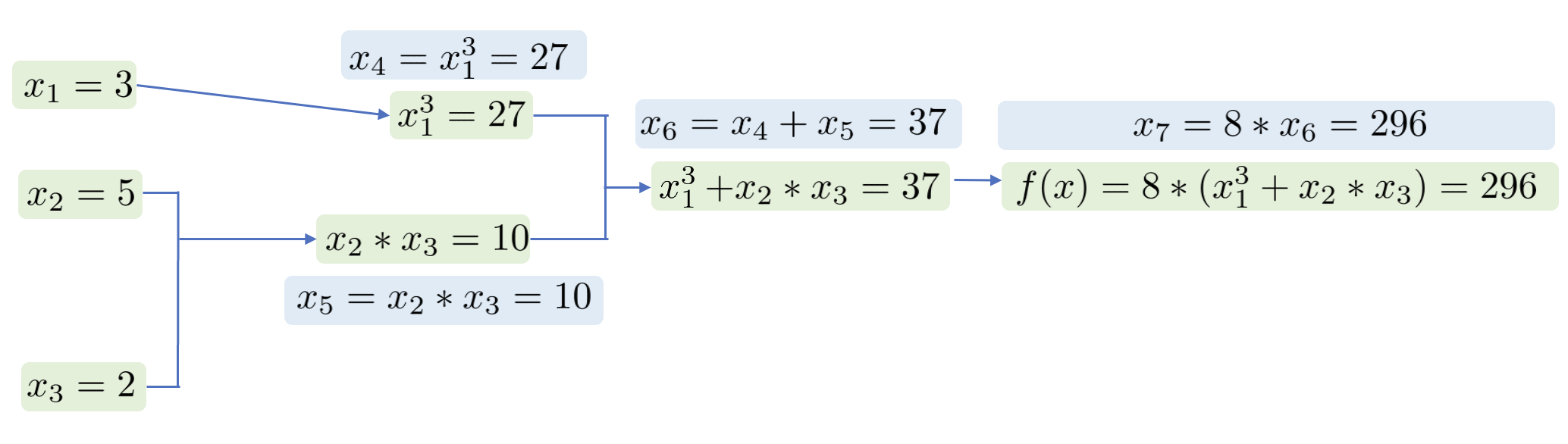}
    \caption{Forward pass for computational graph of f(x).}
    \label{fig:2}
\end{figure}

\begin{figure}[bt]
    \centering
    \includegraphics[width=0.9\textwidth]{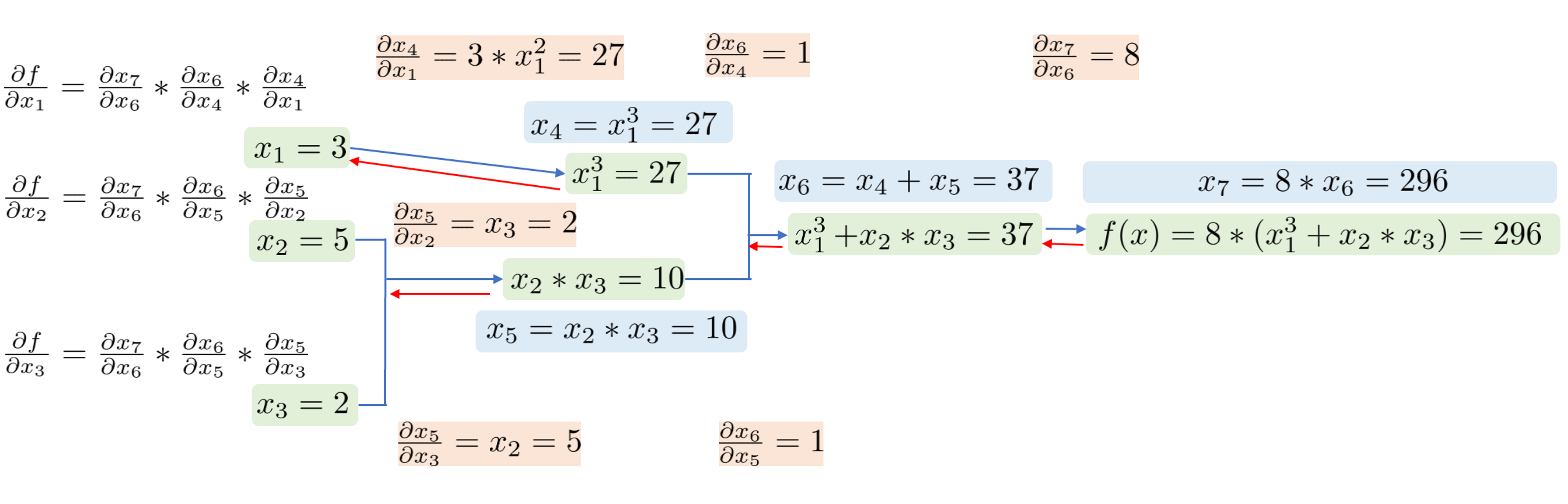}
    \caption{Backward pass for computational graph of f(x).}
    \label{fig:AD}
\end{figure}

Let us consider an example
\begin{equation}
    f(x) = 8(x_{1}^{3}+x_{2}x_{3}),
\end{equation} where $x=(x_{1}, x_{2}, x_{3})$. Now let us consider the computational graph for this function shown in Figure~\ref{fig:CompGraph}.
There are two major steps in the reverse-mode autodiff: forward pass and backward pass. During the forward pass the inputs are passed through the computational graph in a downstream manner as shown in Figure~\ref{fig:2}.
To make it easier we have assigned a new variable for each operation as can be seen in the blue boxes in Figure~\ref{fig:2}. In the backward pass, the chain rule is utilized to calculate the gradient values of the function f(x) with respect to $x_{1},x_{2},x_{3}$ as shown in Figure~\ref{fig:AD}.
Hence by traversing from right to left we calculate the values of $\frac{\partial f}{\partial x}$. Also the partial derivatives are calculated locally, so changing the input points would also change the derivatives. The overall steps can be described as shown below. 

\textbf{Given Function:} 
\begin{equation}
    f(x) = 8 (x_1^3 + x_2 x_3)
\end{equation}
with inputs \( x_1 = 3 \), \( x_2 = 5 \), \( x_3 = 2 \).

\textbf{Forward Pass:} Compute intermediate values:
\begin{align*}
    x_4 &= x_1^3 = 27, \quad x_5 = x_2 x_3 = 10, \\
    x_6 &= x_4 + x_5 = 37, \quad x_7 = 8 x_6 = 296.
\end{align*}

\textbf{Reverse Mode AD:} Compute derivatives using the chain rule:
\begin{align*}
    \frac{\partial x_7}{\partial x_6} &= 8, \quad
    \frac{\partial x_6}{\partial x_4} = 1, \quad
    \frac{\partial x_6}{\partial x_5} = 1, \\
    \frac{\partial x_4}{\partial x_1} &= 3 x_1^2 = 27, \quad
    \frac{\partial x_5}{\partial x_2} = x_3 = 2, \quad
    \frac{\partial x_5}{\partial x_3} = x_2 = 5.
\end{align*}

\textbf{Final Gradients:}
\begin{align*}
    \frac{\partial f}{\partial x_1} &= 8 \times 27 = 216, \quad
    \frac{\partial f}{\partial x_2} = 8 \times 2 = 16, \quad
    \frac{\partial f}{\partial x_3} = 8 \times 5 = 40.
\end{align*}

\textbf{Analytical Derivatives:} 
\begin{align*}
    \frac{\partial f}{\partial x_1} = 8 \times 3x_1^2 = 216, \quad
    \frac{\partial f}{\partial x_2} = 8 \times x_3 = 16, \quad
    \frac{\partial f}{\partial x_3} = 8 \times x_2 = 40.
\end{align*}

\section{Appendix: Experimental Results}
\begin{table}[ht]

\caption{Details of different experimental runs}
\centering
\resizebox{\textwidth}{!}{%
\begin{tabular}{lcccccccccccc}
\toprule
\textbf{File} & \textbf{Power (W)} & \textbf{Flow Rate (L/min)} & \textbf{Amb} & \textbf{Side} & \textbf{Face} & \textbf{In1} & \textbf{In2} & \textbf{Outlet} & \textbf{Inlet} & \textbf{mdot (kg/s)} & \textbf{Delta T (K)} & \textbf{v\_exp} \\
\midrule
A13\_4 & 259.2 & 1.3951 & 22.291 & 23.5721 & 27.4784 & 25.8598 & 25.9011 & 12.5535 & 10.0226 & 0.0233 & 2.5309 & 0.2960 \\
A12\_2 & 259.2 & 2.3431 & 22.2719 & 30.2351 & 34.1676 & 32.5141 & 32.4819 & 21.2002 & 19.7566 & 0.0391 & 1.4436 & 0.4972 \\
A13\_7 & 259.2 & 3.2571 & 22.3062 & 25.1881 & 29.1195 & 27.2510 & 27.2358 & 15.9211 & 14.8075 & 0.0543 & 1.1136 & 0.6912 \\
A11\_1 & 151.8 & 1.3951 & 24.1779 & 22.8727 & 25.2700 & 24.0026 & 23.9805 & 16.4244 & 14.9866 & 0.0233 & 1.4378 & 0.2960 \\
A14\_2 & 151.8 & 3.2571 & 22.2005 & 21.2486 & 23.3733 & 22.4257 & 22.3508 & 15.5734 & 14.9216 & 0.0543 & 0.6518 & 0.6912 \\
A13\_3 & 201.4 & 3.2571 & 22.2347 & 22.9884 & 26.2389 & 24.6879 & 24.5974 & 15.7359 & 14.8670 & 0.0543 & 0.8689 & 0.6912 \\
\bottomrule
\end{tabular}%
}
\end{table}
$C_{p}=4188.5$ $kJ/kg-K$, Density = $999.1$ $kg/m^{3}$

\begin{table}[h!]
\centering
\caption{Case A12\_2 (No data used for training) \hspace{0.2cm} ($Q_{in} = 259.2$ W, $r = 0.005$ mm, $\rho = 999.1 \, kg/m^3$, $C_p = 4188.5 \, J/kg-K$)}
\label{tab:expdata5}
\resizebox{\textwidth}{!}{%
\begin{tabular}{ccccccccccc}
\toprule
\multirow{2}{*}{\textbf{Trial}} & \multirow{2}{*}{\textbf{$h_{nn}$ (W/m$^2$K)}} & \multirow{2}{*}{\textbf{$v_{nn}$ (m/s)}} & \multicolumn{2}{c}{\textbf{Side ($^\circ$C)}} & \multicolumn{2}{c}{\textbf{Face ($^\circ$C)}} & \multicolumn{2}{c}{\textbf{In1 ($^\circ$C)}} & \multicolumn{2}{c}{\textbf{In2 ($^\circ$C)}}  \\
& & & \textbf{Pred} & \textbf{Exp} & \textbf{Pred} & \textbf{Exp} & \textbf{Pred} & \textbf{Exp} & \textbf{Pred} & \textbf{Exp} \\
\midrule
1 & 874.26  & 0.370  & 20.19  & 30.24  & 22.69  & 34.17  & 25.11  & 32.51  & 20.82  & 32.48   \\
2 & 887.08  & 0.389  & 20.27  & 30.24  & 19.75  & 34.17  & 22.20  & 32.51  & 20.17  & 32.48   \\
3 & 912.86  & 0.449  & 20.39  & 30.24  & 23.95  & 34.17  & 25.35  & 32.51  & 23.23  & 32.48   \\
\midrule
\textbf{Mean} & 891.40  & 0.403  & 20.29  & 30.24  & 22.13  & 34.17  & 24.22  & 32.51  & 21.41  & 32.48   \\
\textbf{Std}  & 19.66   & 0.041  & 0.10   & -      & 2.15   & -      & 1.75   & -      & 1.61   & -      \\
\bottomrule
\end{tabular}%
}
\end{table}

\begin{table}[h!]
\centering
\caption{Case A12\_2 \hspace{0.2cm} (Using Face and Side Data, $Q_{in} = 259.2$ W, $r = 0.005$ mm, $\rho = 999.1 \, kg/m^3$, $C_p = 4188.5 \, J/kg-K$)}
\label{tab:expdata6}
\resizebox{\textwidth}{!}{%
\begin{tabular}{ccccccccccc}
\toprule
\multirow{2}{*}{\textbf{Trial}} & \multirow{2}{*}{\textbf{$h_{nn}$ (W/m$^2$K)}} & \multirow{2}{*}{\textbf{$v_{nn}$ (m/s)}} & \multicolumn{2}{c}{\textbf{Side ($^\circ$C)}} & \multicolumn{2}{c}{\textbf{Face ($^\circ$C)}} & \multicolumn{2}{c}{\textbf{In1 ($^\circ$C)}} & \multicolumn{2}{c}{\textbf{In2 ($^\circ$C)}} \\
& & & \textbf{Pred} & \textbf{Exp} & \textbf{Pred} & \textbf{Exp} & \textbf{Pred} & \textbf{Exp} & \textbf{Pred} & \textbf{Exp}  \\
\midrule
1 & 1479.53  & 0.580  & 30.21  & 30.24  & 34.22  & 34.17  & 36.21  & 32.51  & 30.53  & 32.48   \\
2 & 1440.12  & 0.606  & 30.24  & 30.24  & 34.22  & 34.17  & 31.14  & 32.51  & 30.49  & 32.48   \\
3 & 1488.35  & 0.569  & 30.25  & 30.24  & 34.25  & 34.17  & 33.11  & 32.51  & 30.52  & 32.48   \\
\midrule
\textbf{Mean} & 1469.33  & 0.585  & 30.23  & 30.24  & 34.23  & 34.17  & 33.49  & 32.51  & 30.51  & 32.48  \\
\textbf{Std}  & 25.68    & 0.019  & 0.02   & -      & 0.02   & -      & 2.55   & -      & 0.02   & -       \\
\bottomrule
\end{tabular}%
}
\end{table}

\begin{table}[h!]
\centering
\caption{Case A14\_2 \hspace{0.2cm} (Using Face and Side Data, $Q_{in} = 151.8$ W, $r = 0.005$ mm, $C_p = 4188.5 \, J/Kg-K$, $\rho = 999.1 \, Kg/m^3$)}
\label{tab:expdata7}
\resizebox{\textwidth}{!}{%
\begin{tabular}{ccccccccccc}
\toprule
\multirow{2}{*}{\textbf{Trial}} & \multirow{2}{*}{\textbf{$h_{nn}$ (W/m$^2$K)}} & \multirow{2}{*}{\textbf{$v_{nn}$ (m/s)}} & \multicolumn{2}{c}{\textbf{Side ($^\circ$C)}} & \multicolumn{2}{c}{\textbf{Face ($^\circ$C)}} & \multicolumn{2}{c}{\textbf{In1 ($^\circ$C)}} & \multicolumn{2}{c}{\textbf{In2 ($^\circ$C)}}  \\
& & & \textbf{Pred} & \textbf{Exp} & \textbf{Pred} & \textbf{Exp} & \textbf{Pred} & \textbf{Exp} & \textbf{Pred} & \textbf{Exp}  \\
\midrule
1 & 943.48  & 0.736  & 21.24  & 21.25  & 23.44  & 23.37  & 22.09  & 22.43  & 20.65  & 22.35   \\
2 & 1013.64  & 0.756  & 21.24  & 21.25  & 23.44  & 23.37  & 22.04  & 22.43  & 20.68  & 22.35   \\
3 & 1114.24  & 0.720  & 21.24  & 21.25  & 23.44  & 23.37  & 21.97  & 22.43  & 20.73  & 22.35   \\
\midrule
\textbf{Mean} & 1023.79  & 0.737  & 21.24  & 21.25  & 23.44  & 23.37  & 22.03  & 22.43  & 20.68  & 22.35   \\
\textbf{Std}  & 85.83    & 0.018  & 0.002  & -      & 0.003  & -      & 0.062  & -      & 0.040  & -       \\
\bottomrule
\end{tabular}%
}
\end{table}

\newpage

\end{document}